\documentclass[sigconf]{acmart}

\renewcommand\footnotetextcopyrightpermission[1]{} 
\AtBeginDocument{%
  \providecommand\BibTeX{{%
    \normalfont B\kern-0.5em{\scshape i\kern-0.25em b}\kern-0.8em\TeX}}}

\setcopyright{acmcopyright}
\copyrightyear{2018}
\acmYear{2018}
\acmDOI{10.1145/1122445.1122456}

\usepackage{multicol}
\usepackage{multirow}
\usepackage{subcaption}

\usepackage{array}

\begin{document}

\title{Dropout Prediction Uncertainty Estimation \\
Using Neuron Activation Strength}

\author{Haichao Yu}
\authornotemark[1]
\affiliation{UIUC}
\email{haichao3@illinois.edu}
\author{Zhe Chen}
\authornotemark[1]
\affiliation{Google, Inc.}
\email{chenzhe@google.com}
\author{Dong Lin}
\authornote{Authors contributed equally.}
\affiliation{Google, Inc.}
\email{dongl@google.com}
\author{Gil I. Shamir}
\affiliation{Google, Inc.}
\email{gshamir@google.com}
\author{Jie Han}
\affiliation{Google, Inc.}
\email{jieh@google.com}

\newcommand{\eg}{{e.g.}}
\newcommand{\Eg}{{E.g.}}
\newcommand{\ie}{{i.e.}}
\newcommand{\Ie}{{I.e.}}

\newcommand{\movielens}{{MovieLens}}
\newcommand{\mr}{{MovieLens-R}}
\newcommand{\mc}{{MovieLens-C}}
\newcommand{\criteo}{{Criteo}}
\newcommand{\mnist}{{MNIST}}
\newcommand{\emnist}{{EMNIST}}

\newcommand{\PreserveBackslash}[1]{\let\temp=\\#1\let\\=\temp}
\newcolumntype{C}[1]{>{\PreserveBackslash\centering}p{#1}}
\newcolumntype{R}[1]{>{\PreserveBackslash\raggedleft}p{#1}}
\newcolumntype{L}[1]{>{\PreserveBackslash\raggedright}p{#1}}

\newcommand{\uncertainty}{{prediction variance}}

\newcommand{\shirley}{\color{red}}
\newcommand{\dongl}{\color{blue}}
\newcommand{\haichao}[1]{{\color{cyan} #1}}

\begin{abstract}
Dropout has been commonly used to quantify prediction uncertainty, i.e, the variations of model predictions on a given input example. However, using dropout in practice can be expensive as it requires running dropout inferences many times.

In this paper, we study how to estimate dropout prediction uncertainty in a resource-efficient manner. We demonstrate that we can use neuron activation strengths to estimate dropout prediction uncertainty under different dropout settings and on a variety of tasks using three large datasets, \movielens, \criteo, and \emnist. Our approach provides an inference-once method to estimate dropout prediction uncertainty as a cheap auxiliary task. We also demonstrate that using activation features from a subset of the neural network layers can be sufficient to achieve uncertainty estimation performance almost comparable to that of using activation features from all layers, thus reducing resources even further for uncertainty estimation.
\end{abstract}


\maketitle
\pagestyle{plain}

\section{Introduction}
\label{sec:intro}

Uncertainty estimation~\cite{gawlikowski2021survey} is emerging as a very important topic in machine learning 
with a wide range of applications
\cite{yang2015multi,kahn2017uncertainty,feng2018towards,laves2019quantifying,atencia2020uncertainty,wen2020uncertainty}. 
For example, 
in \cite{yang2015multi}, uncertainty sampling was employed to increase the diversity of selected training data. \cite{kahn2017uncertainty} incorporated uncertainty into the cost function for reinforcement learning of autonomous robots to reduce the number of dangerous collisions. \cite{eaton2018towards} examined uncertainty estimation for brain tumor segmentation to calibrate volume estimates.

Ensemble and dropout are two popular methods for {\bf \em prediction uncertainty (PU)}
estimation~\cite{allenzhu21towards,lakshminarayanan2017simple, gal2016dropout}.However, both methods are expensive to deploy in practice. Ensemble requires 
training and deploying multiple copies of a model and running multiple inferences.
These multiple model copies, or ensemble members, produce a set of predictions for a given example and the variation of these predictions can be used to generate prediction uncertainty measurements~\cite{lakshminarayanan2017simple}. 
Unlike ensembles, the dropout approach~\cite{gal2016dropout, laptev2017time} 
trains only one copy of a model using
a given dropout rate $r$.  
During inference time, 
the dropout approach runs multiple inferences with the dropout rate $r$ to 
obtain a collection of predictions from which prediction uncertainty is measured.

Many researchers have studied reducing the cost to estimate
ensemble PU~\cite{valdenegro2019deep,malinin2019ensemble,achrack2020multi,wen2020batchens,chen2021beyond}.
For example, \cite{chen2021beyond} demonstrated that 
{\bf \em neuron activation strength} features 
can be used to infer ensemble PU. Such inference serves as an auxiliary task to the main learning task.
Instead of training multiple copies of the model and using each copy for inference, an auxiliary task is trained to estimate 
the ensemble PU using 
the post neurons activation
values of the target prediction task network. 
To the best of our knowledge,
reducing cost of dropout prediction uncertainty measures still remains an open problem.

\vspace{0.15cm}
{\bf Goal} --- 
We study reducing the inference cost of
estimating {\bf \em dropout prediction uncertainty (DPU)}
using neuron activation strengths. We focus on DPU, 
as it is gaining popularity for a wide variety of real-world 
applications. For example,  
\cite{wen2020uncertainty} used DPU to detect untrustable configurations for molecular simulations to aid quantitative design of materials and devices;
\cite{gal2017deep} used DPU for active learning applications as active learning methods generally rely on uncertainty scores to guide learning and updates of models from small amounts of training data;  
\cite{laves2019quantifying} used DPU to improve computer-aided diagnoses and their robustness for patient safety.

\vspace{0.15cm}
{\bf Challenges} --- 
We address the following challenges.

{\em Various types of data and tasks} ---
DPU has been explored for a wide range of applications. 
Thus, we study DPU estimation for three very different tasks and demonstrate our approach on 
three large datasets, \movielens, \criteo, and \emnist. 

{\em Various training and inference configurations} --- 
We study DPU for 
various configurations and hyperparameters.  For example, one can train the target task with or without dropout, and different dropout rates. 

{\em Activation strength feature selection} ---  
We begin by showing how activation strengths of all neurons from fully-connected layers in the network
can be used to estimate PU.  However, 
using all the activation features may be impractical for very large neural networks \cite{he2016deep,huang2017densely}.
We thus continue by showing that DPU can be estimated even with a 
subset of all the activation features. 

\vspace{0.15cm}
{\bf Our Approach} --- 
In this paper, we propose a multi-task 
learning framework to learn the target prediction
task with an auxiliary task to learn 
DPU simultaneously. During the training time, 
the auxiliary DPU task 
takes neuron activation strengths from the target task model as the input features, and is trained with 
DPU labels, which are the variations of multiple dropout predictions from the target task model. 
As a result, the auxiliary DPU task can be served 
directly as a cost-efficient side task without running 
multiple dropout inferences. We further simplify the 
auxiliary task by reducing its input features 
to only include a subset of the neurons in the network 
from the target task.

We test our approach on regression, binary classification, and multi-class classification target tasks.  These tasks are based on the recommender system benchmark dataset \movielens; 
the Ads click-through rate binary prediction dataset \criteo\ dataset; and the benchmark image digit recognition dataset \emnist. 
We show that our estimation method can be used with different dropout settings on these benchmark datasets.
We observe strong $R^2$ correlations between our method and the true dropout labels in almost all settings.
In particular, our approach achieves strong accuracy (0.8 to 0.9) in identifying whether an example has very high or very low dropout prediction uncertainties.
Furthermore, a substantial reduction of the input size by using only a subset of the neurons also leads to similar results.

{\bf Contributions} --- Our contributions are: 
\begin{itemize}
    \item We propose two configuration setups for
    the DPU estimation task conditioning on 
    whether the target task is trained with dropout
    (Section~\ref{sec:framework}).
    \item We demonstrate that the proposed resource-efficient method provides good estimates of the dropout uncertainty scores 
    on three large public datasets and for both regression and classification tasks (Section~\ref{sec:experiments}).
    \item We show that even a reduced complexity form of our method, using 
    a fraction of neuron activations, still achieves reliable estimates of DPU
    (Section~\ref{sec:experiments}).
\end{itemize}
\section{Related Work}
\label{sec:related_work}

Uncertainty in machine learning models can lead to 
{\bf prediction uncertainties} or model prediction disagreements~\cite{eaton2018towards,fort2019deep,gal2016dropout}.
Some recent work~\cite{anil18large,d2020underspecification,shamir20anti,shamir20smooth,snapp2021synthesizing,summers2021nondeterminism} focused on deep model {\bf irreproducibility} which is caused by nondeterminism in training of deep models.  Designing systems that are able to measure the cumulative prediction uncertainties is, therefore, very important, as these systems can be enhanced to compensate for such uncertainties.

A large volume of uncertainty estimation studies has been under the {\bf Bayesian} umbrella~\cite{paisley2012variational,hoffman2013stochastic, kingma2013auto,blundell2015weight}, where inference also provides an uncertainty score.  Direct Bayesian inference is usually impossible as posterior distributions over model parameters are generally intractable. Some classical Bayesian approaches, such as Markov Chain Monte Carlo, rely on sampling but generally do not scale well with large DNNs~\cite{papamarkou2019challenges}. Instead, approximations, such as variational inference~\cite{paisley2012variational,hoffman2013stochastic, kingma2013auto,blundell2015weight}, can be used. It is not clear, however, how accurately such approximations capture prediction uncertainties. 

{\bf Ensembles} emerged as another class of methods for uncertainty estimation \cite{lakshminarayanan2017simple,valdenegro2019deep,wen2020batchens}.  They are, in fact, a form of Bayesian mixture~\cite{wilson2020bayesian}. An ensemble estimates uncertainty through the distribution of predictions of 
multiple trained copies of the same model.
While ensembles can provide effective PUs, they increase both the training and deployment complexities, and may be impractical in some real large-scale systems.
Researchers have proposed various methods to reduce the computational cost~\cite{valdenegro2019deep,wen2020batchens,mariet2020distilling,chen2021beyond,havasi2020training}.
For example, 
\cite{valdenegro2019deep} proposed to ensemble only the last several layers to 
approximate a deep model ensemble, 
and ~\cite{wen2020batchens} proposed BatchEnsemble to reduce cost by sharing weights among ensemble members.  Such simplifications are usually associated with other cost, such as model complexity. Also, in practical large scale systems, reducing the total number of weights, as proposed by these methods, normally degrades 
inference accuracy.
In \cite{chen2021beyond}, it was proposed to reduce the costs of ensemble deployment by training an auxiliary model on the PUs of multiple trained copies of the model.  This reduced inference costs, but still required the large cost of training ensembles.

{\bf Uncertainty estimation using dropout} is more resource efficient than using ensembles as it does not require training multiple copies of a model. There are abundant studies 
\cite{gal2016dropout, gal2017concrete, kendall2015bayesian}
on dropout uncertainty estimation and its applications. 
For example, 
\cite{gal2016dropout} proposed to use Monte Carlo (MC) dropout to estimate model uncertainty and established a connection to Bayesian inference. 
Further, they extended dropout to work with convolutions~\cite{gal2015bayesian} based on the Bayesian connection.
Concrete dropout ~\cite{gal2017concrete} was introduced to automatically tune the dropout probability in large neural networks, avoiding costly searching for the optimal dropout hyperparameters.
\cite{mukhoti2018evaluating} evaluated MC-dropout and Concrete dropout for uncertainty estimation on semantic segmentation with proposed metrics. \cite{mobiny2021dropconnect} studied MC-DropConnect and MC-Dropout as Bayesian methods in classification for segmentation settings with new uncertainty evaluation metrics.
Further, 
\cite{kendall2015bayesian} presented Bayesian SegNet with MC-Dropout to generate pixel-wise uncertainty estimation and image semantic segmentation tasks.
Unlike the focus of prior work, our focus is on designing a DPU estimation method that avoids the DPU requirement of inferencing multiple times.
\section{DPU Estimation}
\label{sec:framework}
\begin{figure}[t!]
    \includegraphics[width=.75\linewidth]{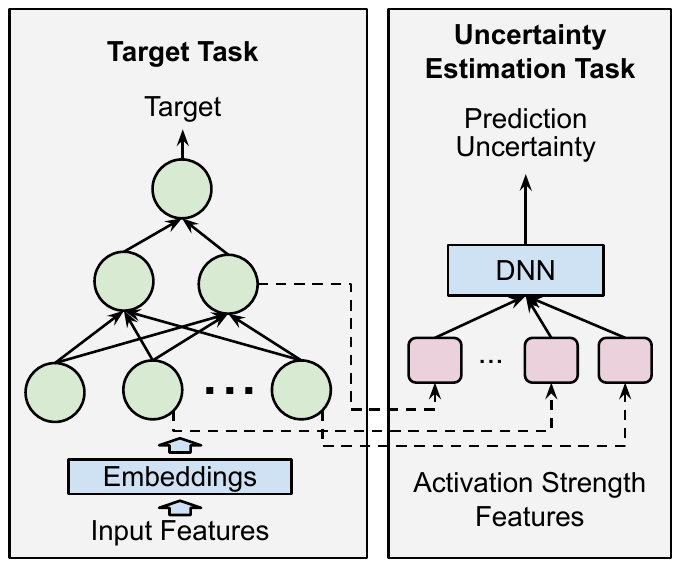}
    \caption{Our framework for prediction uncertainty
    estimation using activation strength.}
    \label{fig:framework}
\end{figure}

Figure~\ref{fig:framework} shows the PU estimation framework.
As shown, it consists of two components: 
the target task and the uncertainty estimation task. 
In this section, we introduce the setup of the two components.

\vspace{0.15cm}
\noindent {\bf Target Task} --- 
The target task represents the original prediction problem, such as 
the rating prediction on \movielens~\cite{harper2015movielens}. 
It takes in the input features and predicts the target objectives. 
For example, on \movielens, the target task takes in 
user and movie features, and 
predicts movie rating.

\begin{figure*}
    \centering
    \begin{subfigure}[t]{0.3\textwidth}
        \centering
        \includegraphics[width=\linewidth]{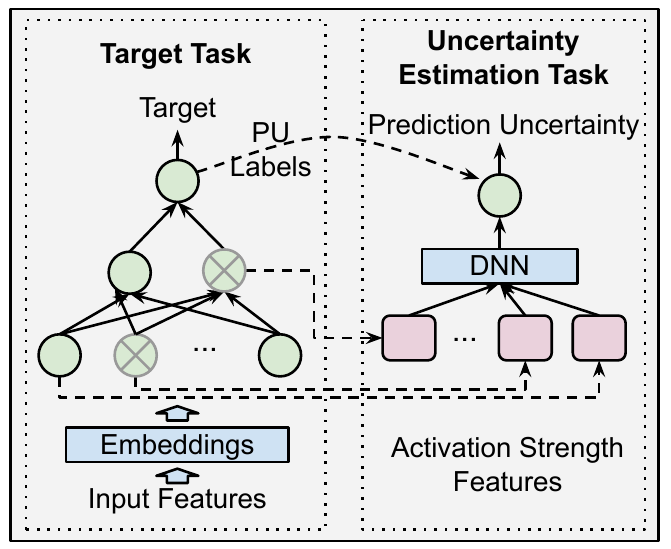}
        \caption{Configuration 1: Dropout enabled on target task.}
    \end{subfigure}
    \hspace{10mm}
    \begin{subfigure}[t]{0.438\textwidth}
        \centering
        \includegraphics[width=\linewidth]{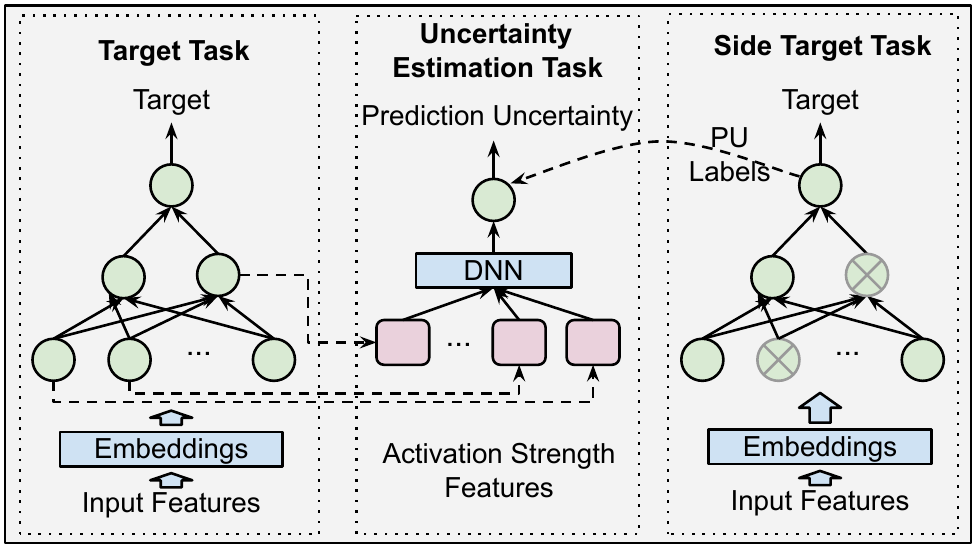}
        \caption{Configuration 2: Dropout not enabled on target task.}
    \end{subfigure}
    \caption{Uncertainty estimation task configuration for dropout predication 
    uncertainties. PU labels represent the prediction uncertainty labels, and 
    the grey crossed neurons represent dropped neurons.}
    \label{fig:configurations}
\end{figure*}
 
\vspace{0.15cm}
\noindent {\bf Uncertainty Estimation Task} --- 
The uncertainty estimation task 
takes in activation strengths from the target task model as its input features.
Those features are concatenated into an input vector that feeds into a DNN 
to estimate the PU for each input example.
DPUs, or training labels for the uncertainty estimation task, are collected by running multiple inferences 
on a target task with dropout enabled.
After the uncertainty estimation model is trained, at the inference time, 
we no longer need to make 
multiple dropout inferences. 
Instead, the model produces DPU estimates from its inputs, the neuron activation strengths. 
The uncertainty estimation task can be seen as  
a cheap auxiliary task alongside the target task. 

\subsection{Prediction Uncertainty Formula}
\label{sec:pv_label_definition}

Following the formula used in~\cite{lakshminarayanan2017simple,gal2016dropout,chen2021beyond}, DPUs used in this paper are defined in the following.
Assume running dropout inferences
$N$ times produces $N$ predictions  $\{\hat{y}_1(x), ..., \hat{y}_N(x)\}$
for each example $x \in \{x\}$, then: 

{\bf For binary classification and regression}:
$PU(x)$ is just the prediction standard deviation 
\begin{equation}
    PU(x) \stackrel{\triangle}{=} \sqrt{\frac{\sum_{i=1}^{N} (\hat{y}_i(x) - \bar{y}(x))^2}{N-1}},
    \label{equ:pv_std}
\end{equation}
where $\bar{y}(x) = \frac{1}{N}\sum_{i=1}^{N} \hat{y}_i(x)$; 

{\bf For multi-class classification}: each $\hat{y}_i(x)$ is a vector  of probabilities. The vector's length is $K$ if there are $K$ classes. Therefore,
a KL divergence based metric is used. Assume the vector is denoted as $\hat{\vec{p}}_i(y|x)$:
\begin{equation}
    PU(x) \stackrel{\triangle}{=} \sum_{i=1}^{N} KL(\hat{\vec{p}}_i(y|x)\ ||\ \bar{\vec{p}}(y|x)),
    \label{equ:pv_kl}
\end{equation}
where $\bar{\vec{p}}(y|x) = \frac{1}{N}\sum_{i=1}^{N} \hat{\vec{p}}_i(y|x)$ is 
the averaged probability vector over the $N$ runs.

\subsection{Uncertainty Estimation Task Configurations}
\label{sec:configurations}

As shown in Figure~\ref{fig:configurations}, we consider two real-world configurations for the DPU estimation task depending on whether the deployed target task is trained with dropout. 

\begin{itemize}
\item{\bf Configuration 1} describes a framework in which the deployed target task is trained with dropout enabled. 
At target-task inference time, dropout is disabled. Neuron activations are inputted to the 
uncertainty estimation task. PU labels are collected by running target task predictions multiple times with dropout enabled for each target-task input.

\item{\bf Configuration 2} describes a framework in which the deployed target task is trained without dropout. 
Many real-world production models avoid dropout in order to 
achieve better accuracy~\cite{covington2016deep}. 
Again, neuron activations are inputted to the uncertainty estimation task. 
A side target task is setup with dropout to gather DPUs by generating 
multiple predictions for a given example during training of the auxiliary task. The side target task can be discarded after the uncertainty estimation model is trained.
Inference in deployment uses only the target task model without dropout and the trained uncertainty estimation model.
\end{itemize}

Target tasks are run until convergence.
Follow the most typical dropout setup, we apply dropout only to the fully-connected layers~\cite{gal2016dropout, laptev2017time}.  In particular, we neither enable dropout on convolution layers on \emnist, nor on embeddings~\cite{yao2020self}. 

\subsection{Neuron Activations}
\label{sec:features}
In this paper, ReLU~\cite{nair2010rectified} activation function is used.
Activation outputs are inputted to the uncertainty estimation task. We construct input features from activation outputs in two ways: 1) as binary features to indicate whether neurons are activated or not;  2) as normalized real-valued features.

The most straightforward approach is to use all activations as inputs to the uncertainty estimation task.  However, this can be impractical and under-perform due to limited training data and overfitting. For example,
the \emnist\ target task model has layers of widths [3456, 1024, 120, 84], with 3456 and 1024 denoting two convolutional layer widths, and 120 and 84 denoting two fully-connected layer width, resulting 4684 neurons in total.  The uncertainty estimation task with an initial hidden layer of width 100 will have 468K hidden link weights for just this first layer.  The whole \emnist\ dataset consists of only 280K examples, thus the uncertainty estimation task will be highly overparameterized.  Therefore, it may be better to use activations from a subset of the hidden layers only.  This reduces model complexity as well as mitigates potential overfitting. 
Specifically, we observe that uncertainty estimation task performs worse on validation data if using neurons from all the layers of the target task on the \emnist\ dataset than using only the two fully-connected hidden layers (FCLs). We thus limit experiments in this paper to use activation inputs only from the FCLs to avoid overparameterization.

In addition, we want to further
reduce the complexity and cost of uncertainty estimation task by
selecting a subset of FCL activation strengths. As different layers
in a DNN reflect different representations of input data~\cite{zhang2019all, tishby2015deep}, 
it might be possible to select the subset of neurons by layers. We demonstrate our findings in Section~\ref{sec:pruning}.
\section{Experiment Setup}
\label{sec:experiment_setup}

\subsection{Datasets}
We consider three datasets to evaluate the proposed framework:
\begin{itemize}
    \item \movielens~\cite{harper2015movielens}:
    The \movielens\ 1M dataset\footnote{
    \url{http://files.grouplens.org/datasets/movielens/ml-1m-README.txt}} 
    is a benchmark dataset for evaluation of recommender systems. 
    It features the task of using user and movie related features to predict movie ratings. 
    This dataset contains 1M movie ratings of 4000 movies from 6000 users.

    \item \criteo: 
    The Criteo Display Advertising challenge ~\footnote{https://www.kaggle.com/c/criteo-display-ad-challenge} features a binary classification task to predict Ads click-through rate. Labels are 1 for clicked examples, and 0, otherwise. This dataset has ~40M examples.

    \item \emnist:
     The \emnist-Digits~\cite{cohen2017emnist} dataset consists of a set of 
     280K handwritten images for a digit recognition task, 
     which is an extension of the 70K \mnist~\cite{lecun1998gradient} dataset. 
\end{itemize}

\vspace{0.15cm}
\noindent {\bf Training/Testing Data} --- 
We set up the training and testing data for the target task 
and the uncertainty estimation task by splitting the data 
into training $D_{train}$ and testing $D_{test}$ sets.
\emnist\ is split randomly into two equal sets, to ensure there are sufficient (140K) test examples.
\movielens\ is split to 60\% training and 40\% test examples.
For \criteo, we use the standard splitting~\cite{ovadia2019can}, with
37M training examples and 4.4M validation and test examples. 
The test sets are further split randomly into two equal parts 
$D'_{train}$ and $D'_{test}$ for training and testing the uncertainty estimation task.

\subsection{Target Tasks Setup}
\label{sec:target_tasks}

We consider four target tasks on the three datasets.
For \movielens\ and \criteo, the target task model contains a trained
embedding layer connected to input features.
For \emnist, embeddings are not necessary. 

\vspace{0.1cm}
\noindent {\bf \movielens\ Regression (\movielens-R)} --- 
The target task takes in 
user features (\ie, identification, gender, age, and occupation) 
and movie features (\ie, identification, title and genres), and 
predicts movie ratings as a regression task. The movie ratings are integers from 1 to 5.
We use Mean Squared Error (MSE) as the loss function.

Each model is constructed with 3 fully-connected layers (FCL) of widths [250, 100, 50].
Each model trains for 20 epochs to converge. 
We use user ID and item ID embedding widths of 8~\cite{he2017neural}, 
user age embedding of width 3, user 
occupation embedding of width 5, and gender and 
genre features of one-hot encoding. 
We do not use the title textual feature for simplicity.
Therefore, the total input embedding layer width is 43.

\vspace{0.1cm}
\noindent {\bf \movielens\ Classification (\movielens-C)} --- The same model as the one for the regression task is used here, 
except for the prediction objective.
Instead of regression, the rating is treated as a multi-class classification to 5 integer value ratings (from 1 to 5).  The model is thus trained to minimize Softmax cross-entropy loss.

\vspace{0.1cm}
\noindent {\bf \criteo} --- This target task uses 13 numerical and 26 categorical 
features that determine click-through-rate.  High valency categorical features are encoded into embedding vectors, and low valency ones are hard encoded into one-hot vectors, taking a nonzero binary value for the category that is present.  The overall input specifications follow those in \cite{ovadia2019can}. 
The inputs feed into an MLP with layer sizes [250, 100, 50], which are trained with cross-entropy loss on $\{0,1\}$ labels (1 denoting a click). The model output is mapped to a click probability using the Sigmoid function. The model is trained for one epoch with each example visited once.

\vspace{0.1cm}
\noindent {\bf \emnist} --- 
This digit recognition task takes a $28 \times 28$ pixels grayscale image representing a digit from 0 to 9, and 
predicts the digit in the image. We used the LeNet-5~\cite{lecun1998gradient} 
10-category classification model.
The model is composed of two convolutional layers with feature map sizes $[3456, 1024]$, 
each followed by a max-pooling layer, and two FCLs of sizes $[120, 84]$. The model is trained for 50 epochs to converge.  

All the above target tasks use Batch Normalization~\cite{ioffe2015batch}. 
A batch size of 1024 is used for \movielens\ and \criteo,
and a batch of 128 is used on \emnist~\cite{wan2013regularization}.
We use Adam optimizer with a initial learning rate 0.01 for 
\criteo, 0.1 for \emnist, and 0.001 for \mc\ and \mr. 

\subsection{Uncertainty Estimation Task Setup}

As pointed out in Section~\ref{sec:configurations}, we use two configurations for the uncertainty estimation task. The dataset slice $D'_{train}$ is used to train the uncertainty estimation task and we use a 2-layer MLP with layer widths of [100, 50].
Inputs to the task are either all activation strengths of the fully-connected hidden layers of the target task, or activation strengths from subsets of the layers.  Inputs can consist of activation values, and/or their binary indicators. 
We collect the DPUs, i.e. labels, by running dropout inference on 
the target task model or the side target task model 100 times. All the models are trained for 50 epochs. We use the Adam optimizer with a initial learning rate 0.001, decaying by 0.1 at the 30-th and 40-th epochs. 
Dataset slice $D'_{test}$ is used to test the uncertainty estimation performance. We repeat each uncertainty estimation task 20 times and report the average performance.

\vspace{0.15cm}
\noindent{\bf Task Objectives} --- 
The prediction uncertainty estimation task is set up as either a
regression or a classification task: 

\begin{itemize}
\item {\bf Regression} ---
    The model directly estimates the prediction uncertainty using MSE as the loss function. By directly optimizing MSE, the output could have a range that's too wide.  To mitigate this, we either clip or transform the outputs.  With clipping, outputs are capped within the minimum and maximum measured prediction uncertainty values.
    For \emnist\ (as discussed in Section~\ref{sec:dropout_estimation_performance}),
    we transform raw prediction and labels to log scale.
\item {\bf Classification} --- 
    For classification, prediction uncertainty is uniformly divided by frequency
    into multiple buckets, and we predict the uncertainty bucket.  We use 5 buckets
    and cross-entropy loss.
    Bucket 0 is the most certain class and 
    4 the most uncertain one. 
\end{itemize}

\section{Experiments}
\label{sec:experiments}

We report empirical results below.  The results demonstrate that DPUs can be estimated with a DNN model that uses activation strengths as input.  The results hold when we use all neurons of FCLs or a subset of neurons of the FCLs as input features to the uncertainty estimation task. 

\begin{figure*}[t]
    \centering
    \begin{subfigure}[t]{0.245\textwidth}
        \centering
        \includegraphics[width=\linewidth]{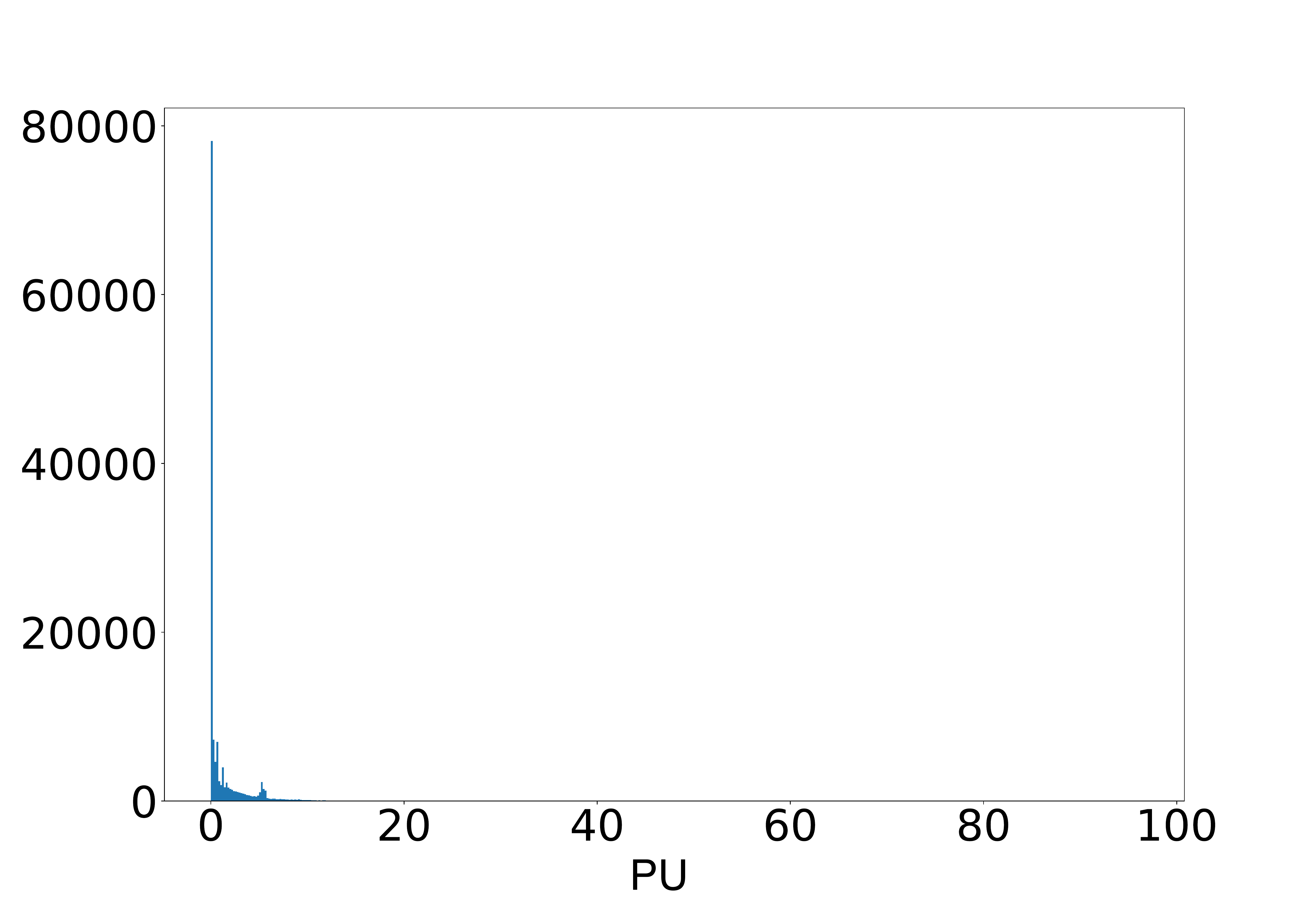}
        \caption{\emnist}
    \end{subfigure}
    \begin{subfigure}[t]{0.245\textwidth}
        \centering
        \includegraphics[width=\linewidth]{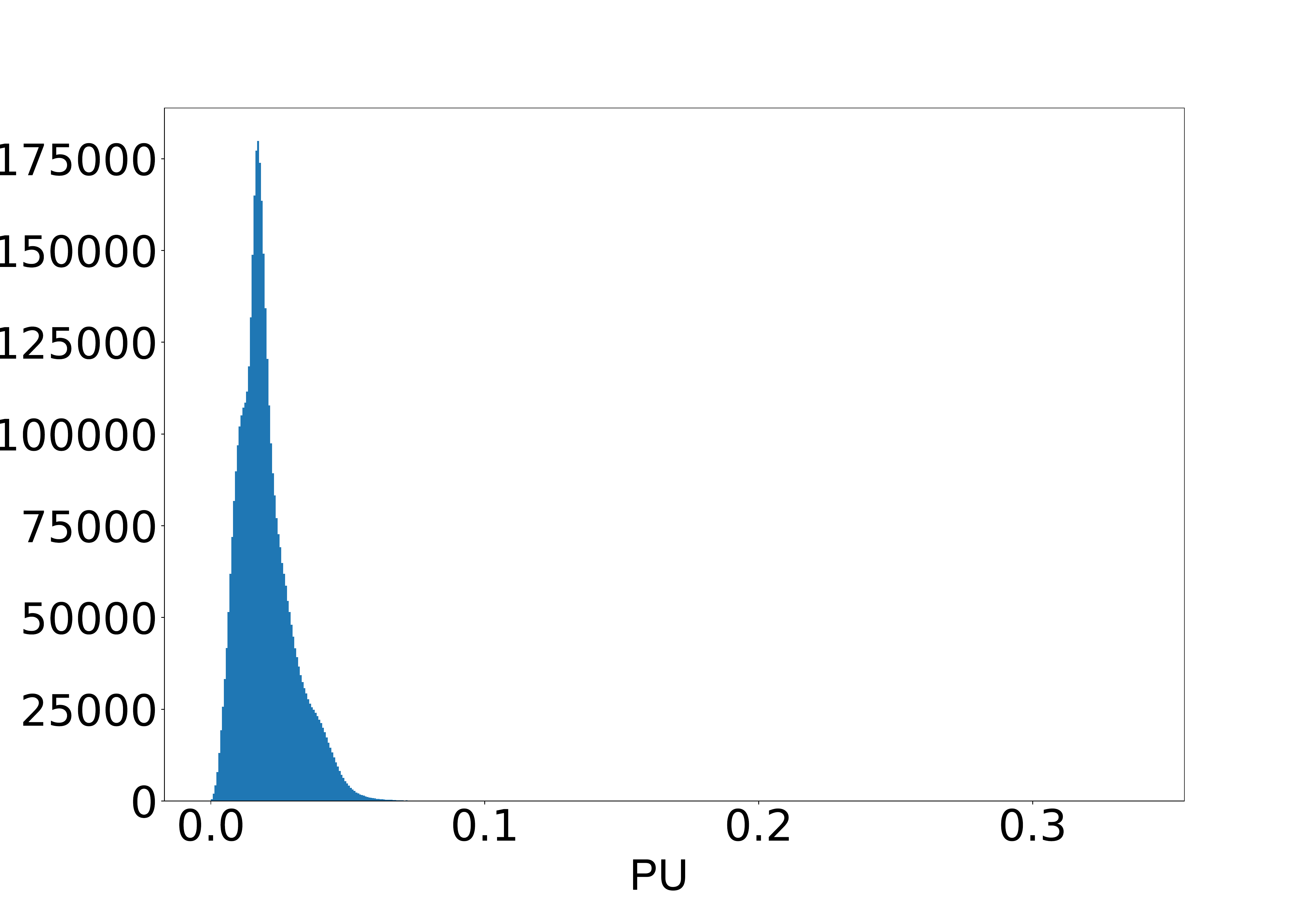}
        \caption{\criteo}
    \end{subfigure}
    \begin{subfigure}[t]{0.245\textwidth}
        \centering
        \includegraphics[width=\linewidth]{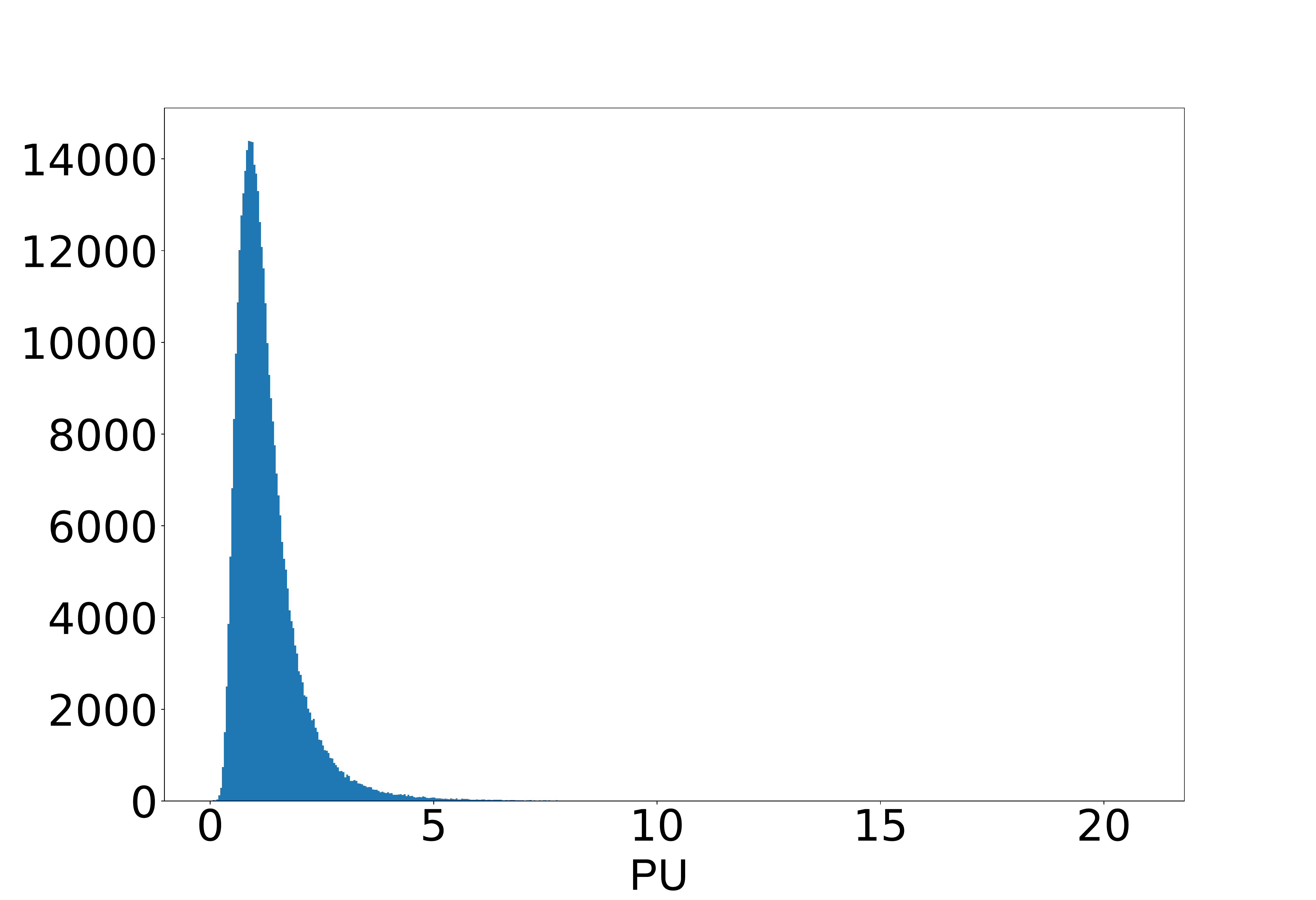}
        \caption{\mc}
    \end{subfigure}
    \begin{subfigure}[t]{0.245\textwidth}
        \centering
        \includegraphics[width=\linewidth]{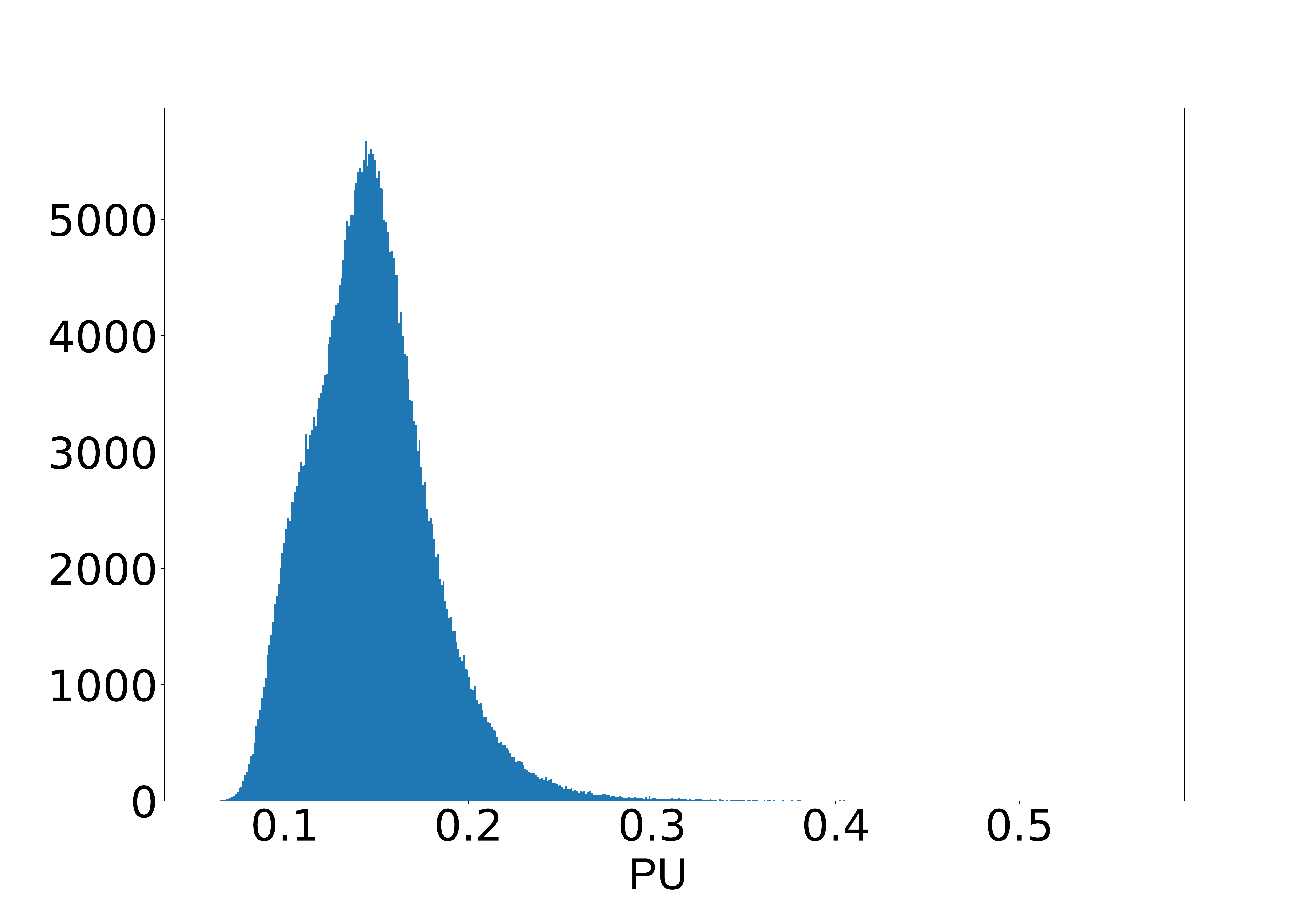}
        \caption{\mr}
    \end{subfigure}
    \caption{Dropout Prediction uncertainty histograms of different datasets when dropout rate is 0.1.}
    \label{fig:uncer_distribution}
\end{figure*}

\begin{figure*}[t]
    \centering
    \begin{subfigure}[t]{0.245\textwidth}
        \centering
        \includegraphics[width=\linewidth]{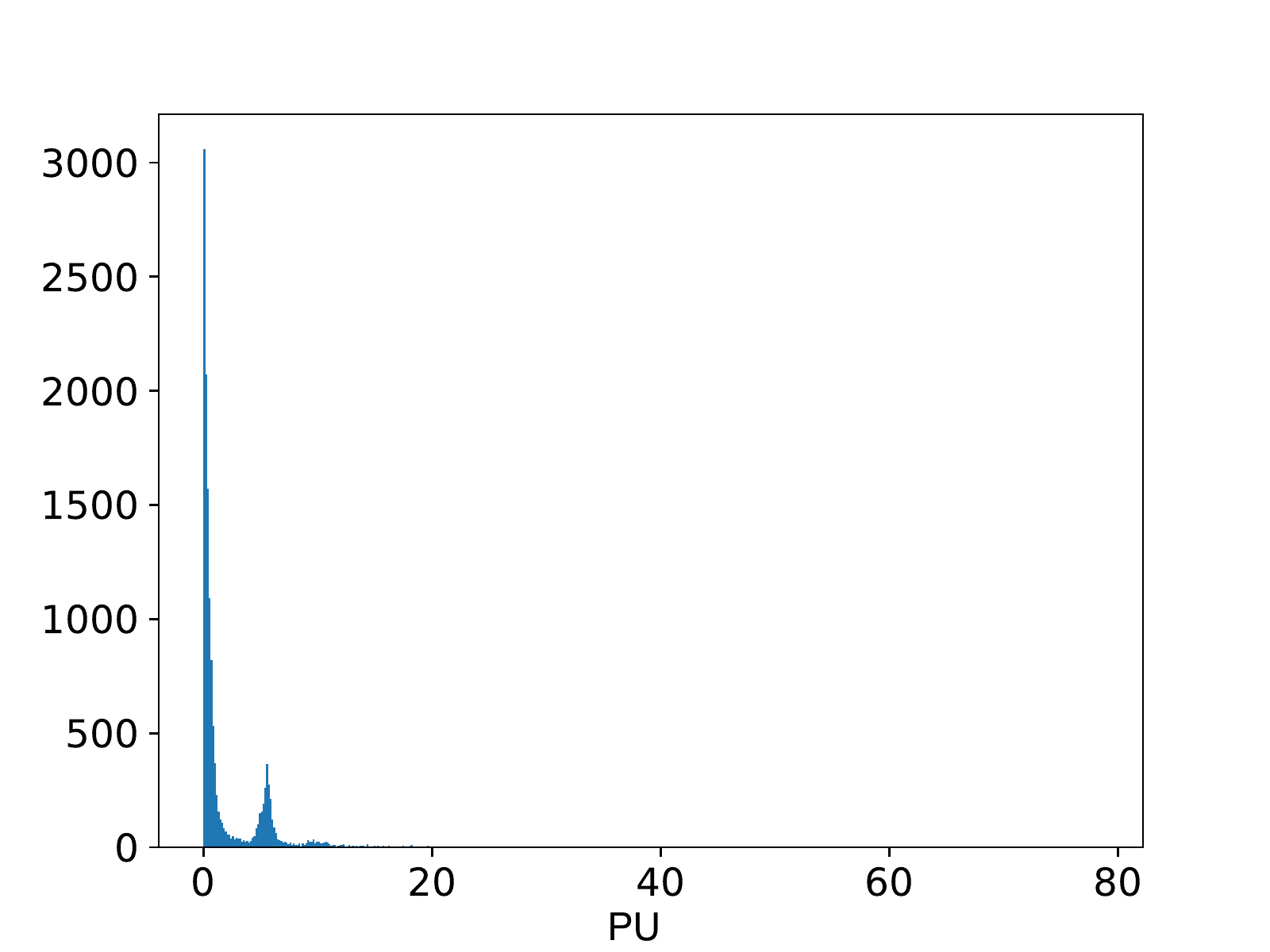}
        \caption{Digit 0 (dropout 0.1)}
    \end{subfigure}
    \begin{subfigure}[t]{0.245\textwidth}
        \centering
        \includegraphics[width=\linewidth]{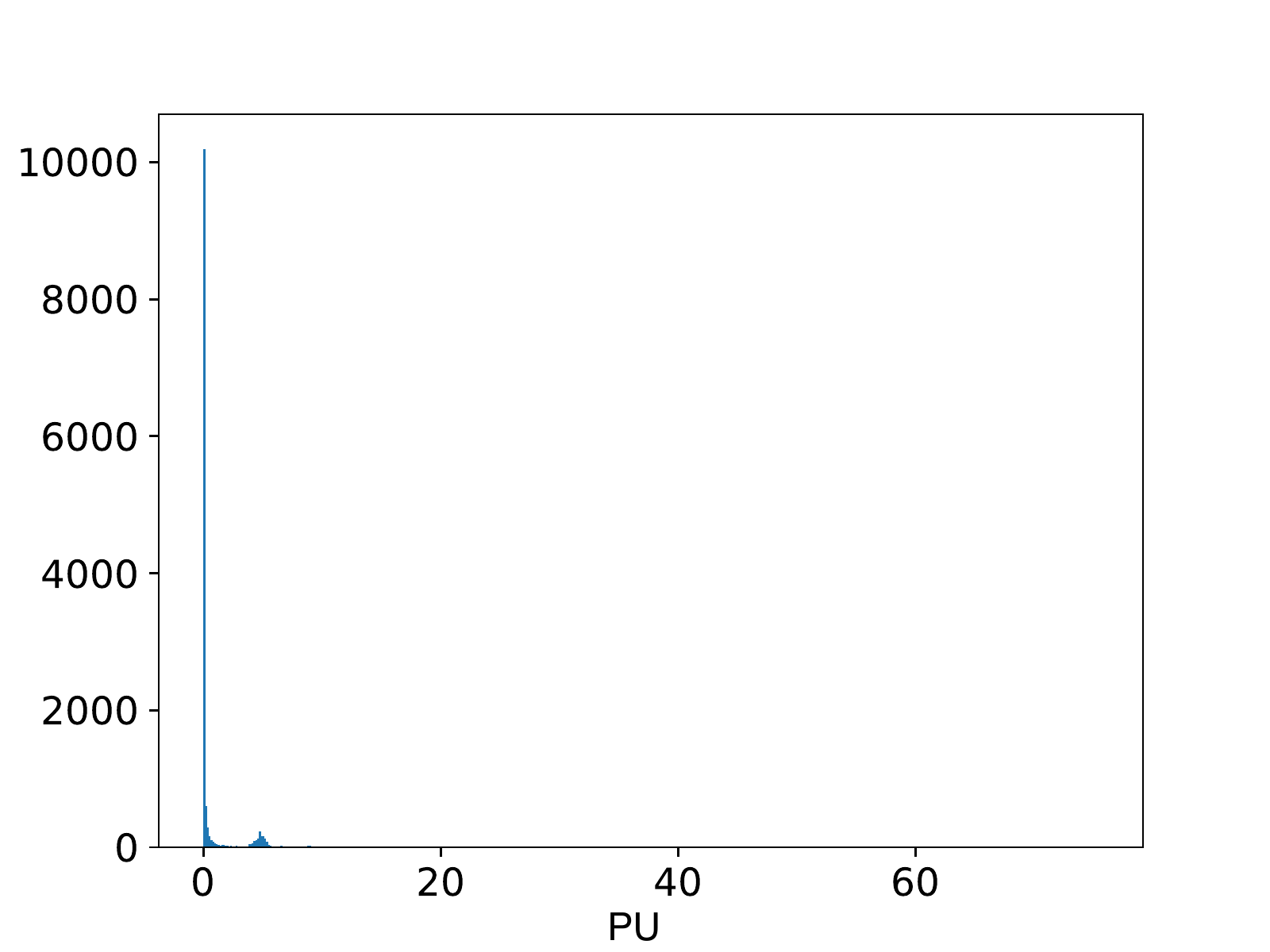}
        \caption{Digit 4 (dropout 0.1)}
    \end{subfigure}
    \begin{subfigure}[t]{0.245\textwidth}
        \centering
        \includegraphics[width=\linewidth]{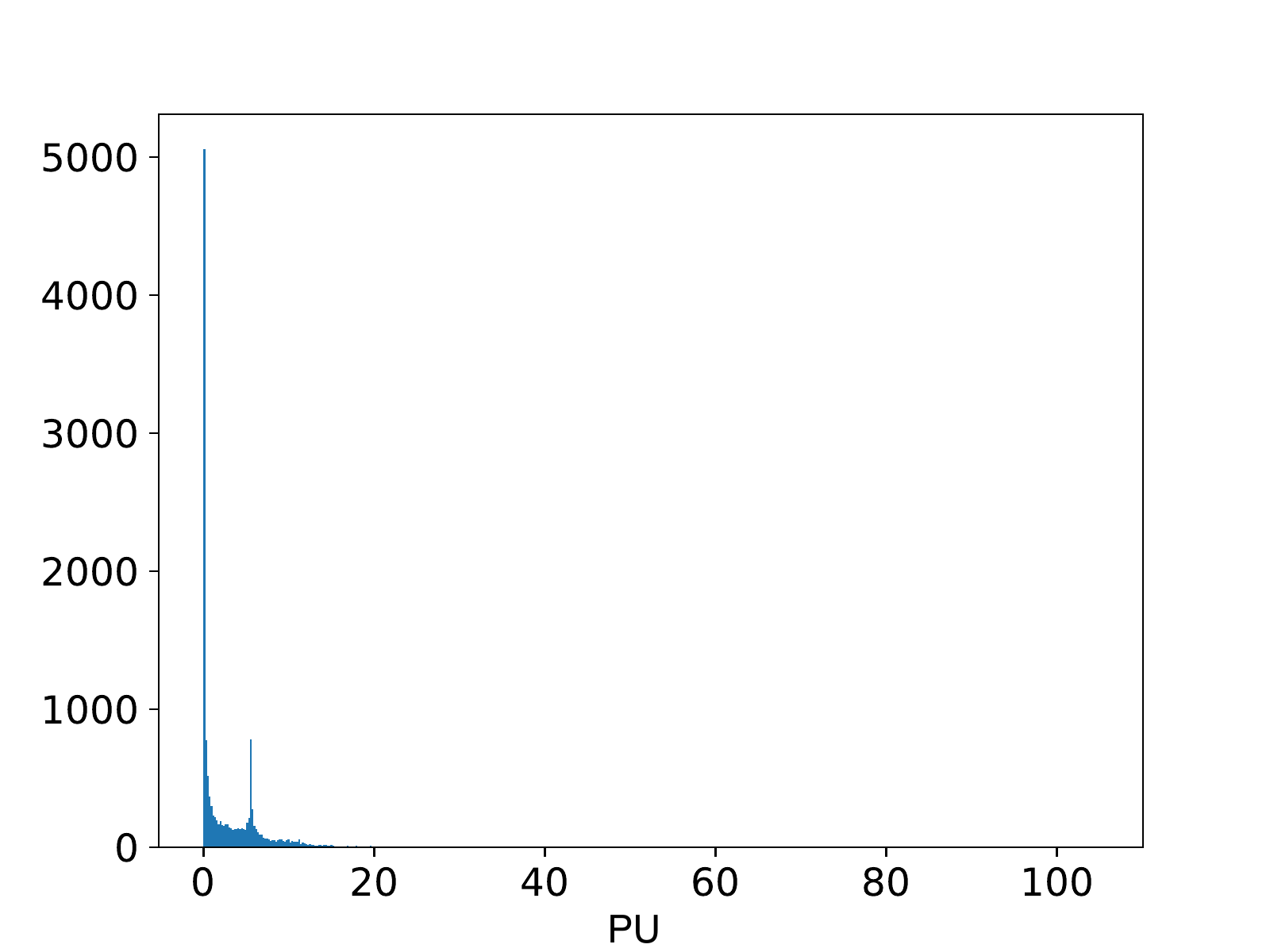}
        \caption{Digit 0 (dropout 0.3)}
    \end{subfigure}
    \begin{subfigure}[t]{0.245\textwidth}
        \centering
        \includegraphics[width=\linewidth]{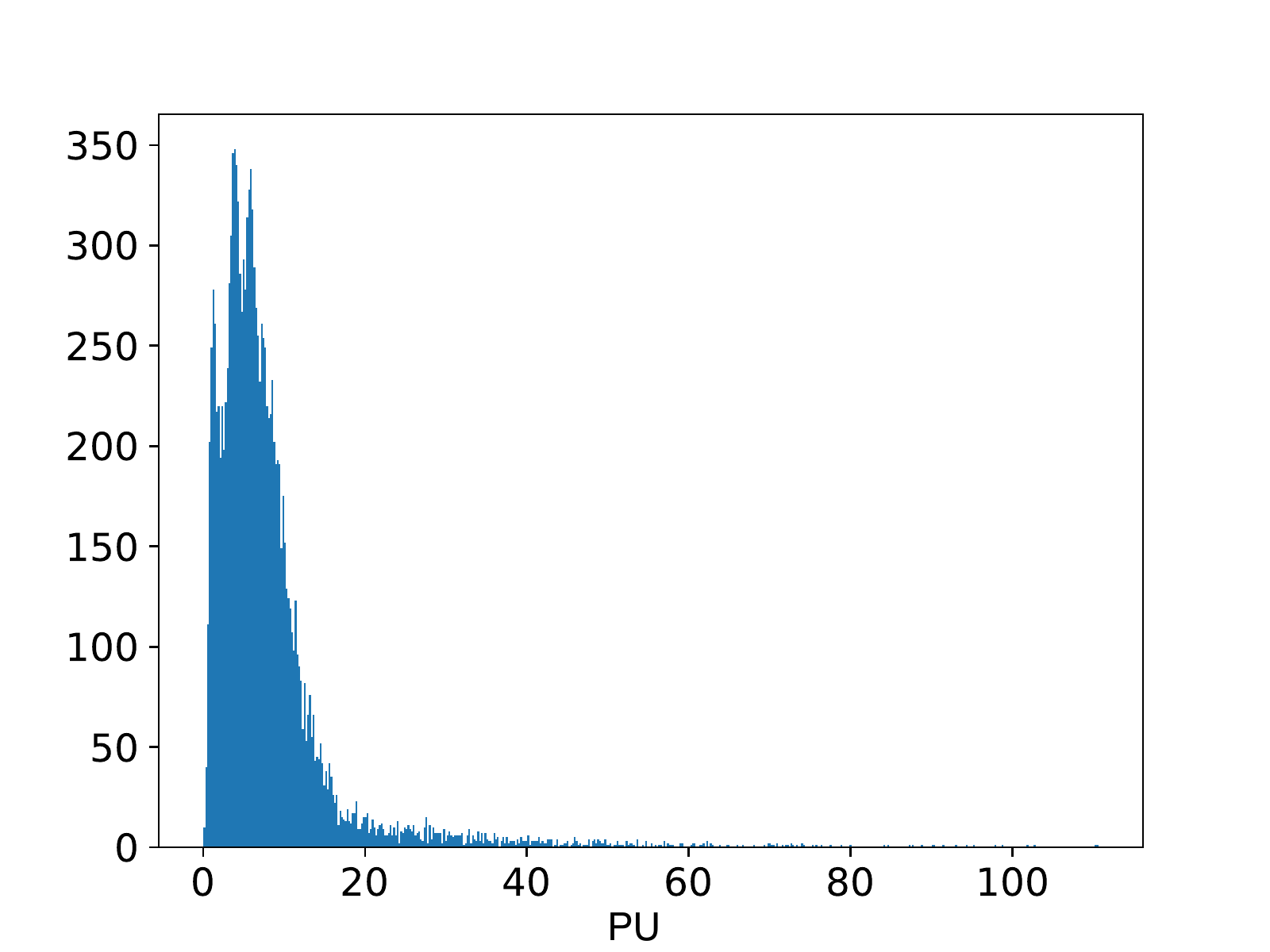}
        \caption{Digit 4 (dropout 0.3)}
        \label{fig:digit_4}
    \end{subfigure}
    \caption{Dropout prediction uncertainty histograms of digits 0 and 4 in \emnist\ for dropout rate 0.1 and 0.3. 
    }
    \label{fig:emnist_digitwise}
\end{figure*}

\begin{table}[t]
    \centering
    \footnotesize
    \setlength{\tabcolsep}{2pt}
    \begin{tabular}{l|c|C{0.63cm}C{0.63cm}C{0.63cm}|C{0.63cm}C{0.63cm}C{0.63cm}}
        \hline
         \multirow{2}{*}{Target Tasks} & 
         \multirow{2}{*}{Ensemble} & \multicolumn{3}{c|}{Inference with Dropout} 
         & \multicolumn{3}{c}{Inference w/o Dropout} \\ \cline{3-8}
         \tiny
         & & 0.1 & 0.3 & 0.5
         & 0.1 & 0.3 & 0.5 \\ 
        
        \hline
        \emnist\ accuracy & 0.992 & 0.990 & 0.982 & 0.962 & 0.993 & 0.992 & 0.991 \\
        \criteo\ AUC & 0.799 & 0.797 & 0.793 & 0.787 & 0.799 & 0.798 & 0.795 \\
        \mc\ accuracy    & 0.471 & 0.469 & 0.467 & 0.459 & 0.472 & 0.473 & 0.470 \\
        \mr\ MSE    & 0.777 & 0.799 & 0.802 & 0.824 & 0.775 & 0.776 & 0.780 \\
        \hline
    \end{tabular}
    \caption{Target task performance for different dropout rates
    with dropout enabled in training, and with and without dropout enabled in inference. 
    }
    \label{tab:original_perform}
\end{table}

\begin{table}[t]
    \footnotesize
    \setlength{\tabcolsep}{3pt}
    \centering
    \begin{tabular}{ll|cc|cc|cc}
        \hline
        \multicolumn{2}{c|}{\multirow{2}{*}{Target Tasks}}  & \multicolumn{2}{c|}{Ensemble} & \multicolumn{2}{c|}{0.1} & \multicolumn{2}{c}{0.3}  \\
        \cline{3-8}
        & & mean & SD & mean & SD & mean & SD \\
        \hline
        \multirow{2}{*}{Classification} & \multicolumn{1}{|l|}{\emnist} & 1.897 & 8.932 & 1.693 & 5.555 & 5.702 & 8.950 \\
        & \multicolumn{1}{|l|}{\mc}     & 4.488 & 3.096 & 1.265 & 0.758 & 2.113 & 1.111\\
        \hline
        \multirow{2}{*}{Regression}& \multicolumn{1}{|l|}{\criteo} & 0.027 & 0.015 & 0.02 & 0.011 & 0.032 & 0.014 \\
        & \multicolumn{1}{|l|}{\mr}     & 0.179 & 0.058 & 0.146 & 0.034 & 0.161 & 0.055 \\
        \hline
    \end{tabular}
    \caption{The mean and standard deviation of DPU for different dropout rates.}
    \label{tab:uncer_range}
\end{table}

\begin{table}[t]
    \centering
    \footnotesize
    \setlength{\tabcolsep}{2pt}
    \begin{tabular}{l|ccccc|ccccc}
        \hline
        \multirow{2}{*}{Target Tasks} 
        & \multicolumn{5}{c|}{Configuration 1} 
        & \multicolumn{5}{c}{Configuration 2} \\ \cline{2-11}
        & 0.1 & 0.2 & 0.3 & 0.4 & 0.5 
        & 0.1 & 0.2 & 0.3 & 0.4 & 0.5 \\
        \hline
        \emnist & 0.513 & 0.574 & 0.554 & 0.542 & 0.456 
        & 0.450 & 0.494 & 0.445 & 0.441 & 0.307 \\
        \criteo & 0.935 & 0.944 & 0.946 & 0.944 & 0.942 
        & 0.839 & 0.853 & 0.859 & 0.862 & 0.850 \\
        \mc & 0.879 & 0.900 & 0.912 & 0.922 & 0.925 
        & 0.477 & 0.545 & 0.595 & 0.640 & 0.667 \\
        \mr & 0.853 & 0.900 & 0.931 & 0.945 & 0.949 
        & 0.575 & 0.630 & 0.722 & 0.753 & 0.771 \\
        \hline
    \end{tabular}
    \caption{Regression $R^2$ of DPU estimation for different configurations and dropout rates. }
    \label{tab:dropout_results_regression}
\end{table}

\begin{table}[t]
    \centering
    \footnotesize
    \setlength{\tabcolsep}{2pt}
    \begin{tabular}{l|ccccc|ccccc}
        \hline
        \multirow{2}{*}{Target Tasks} 
        & \multicolumn{5}{c|}{Configuration 1} 
        & \multicolumn{5}{c}{Configuration 2} \\ \cline{2-11}
        & 0.1 & 0.2 & 0.3 & 0.4 & 0.5 
        & 0.1 & 0.2 & 0.3 & 0.4 & 0.5 \\
        \hline
        \emnist & 0.530 & 0.497 & 0.463 & 0.429 & 0.463 
        & 0.471 & 0.447 & 0.418 & 0.384 & 0.411 \\
        \criteo & 0.808 & 0.793 & 0.769 & 0.731 & 0.689
        & 0.714 & 0.703 & 0.686 & 0.656 & 0.619 \\
        \mc & 0.639 & 0.660 & 0.674 & 0.688 & 0.690
        & 0.442 & 0.450 & 0.458 & 0.467 & 0.473 \\
        \mr & 0.604 & 0.646 & 0.670 & 0.683 & 0.695
        & 0.494 & 0.503 & 0.511 & 0.518 & 0.521 \\
        \hline
    \end{tabular}
    \caption{Classification accuracy of DPU estimation for different configurations and dropout rates. }
    \label{tab:dropout_results_classification}
\end{table}

\begin{figure*}[th]
    \centering
    \begin{subfigure}[t]{0.245\textwidth}
        \centering
        \includegraphics[width=\linewidth]{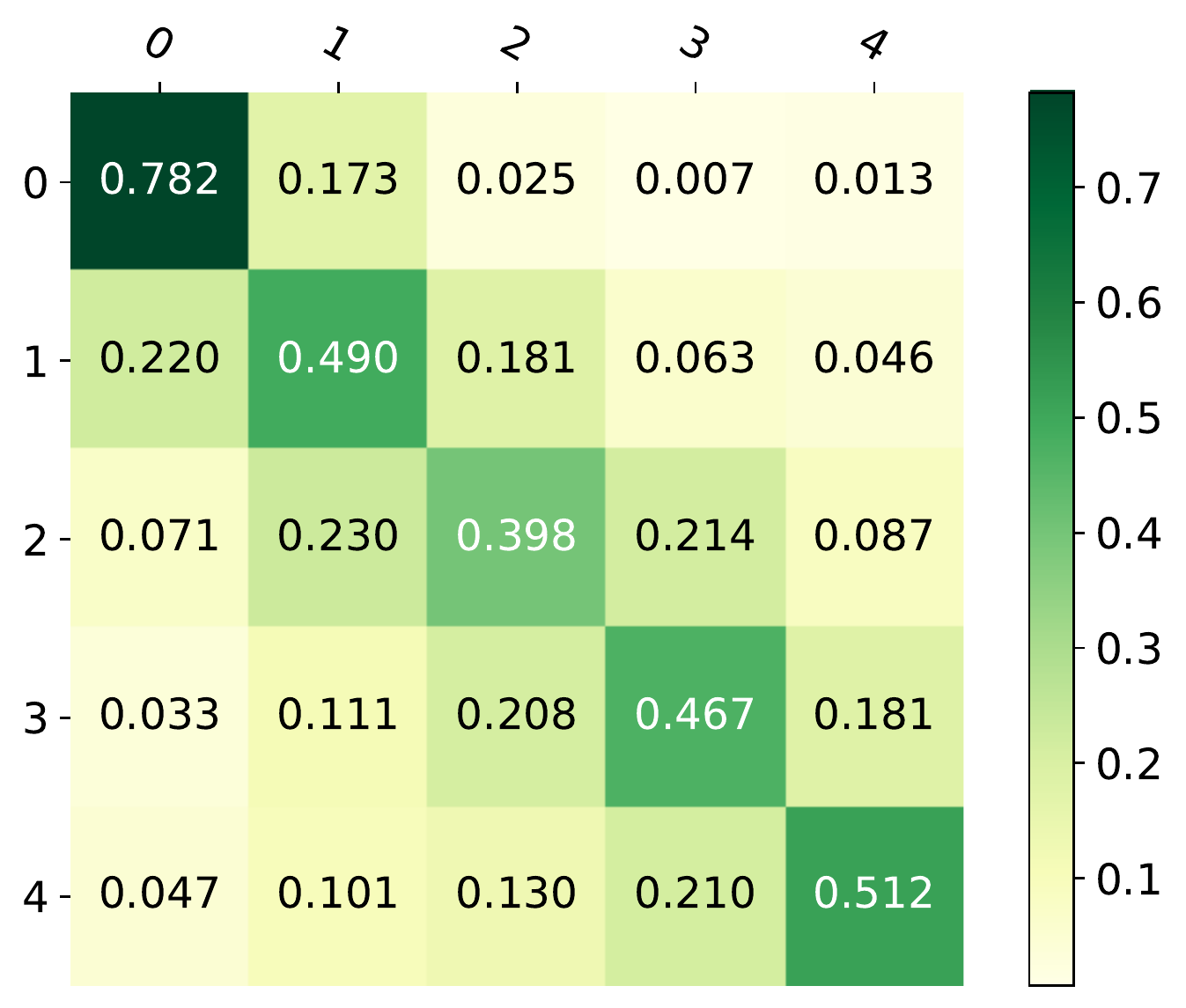}
        \caption{\emnist}
    \end{subfigure}
    \begin{subfigure}[t]{0.245\textwidth}
        \centering
        \includegraphics[width=\linewidth]{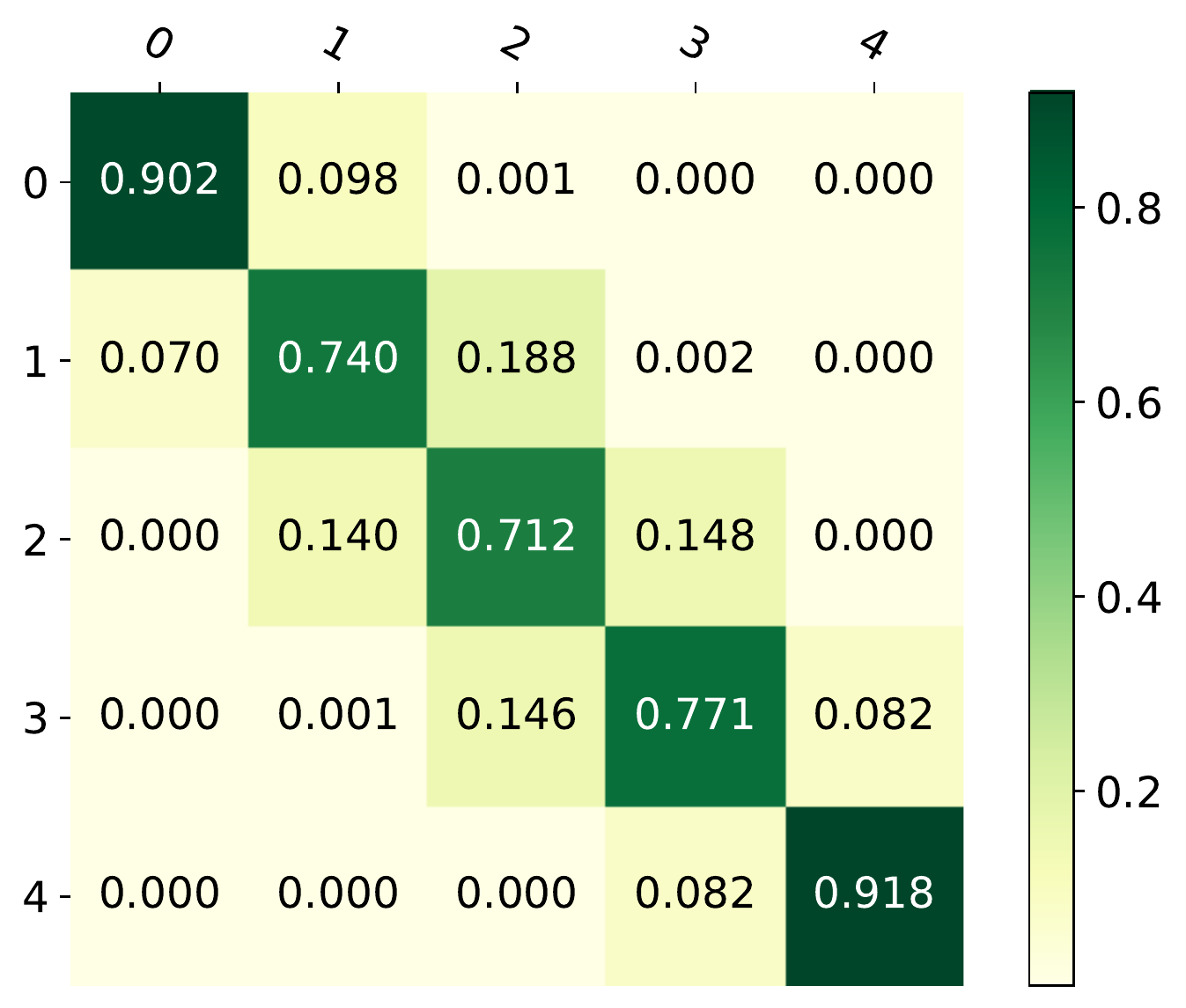}
        \caption{\criteo}
    \end{subfigure}
    \begin{subfigure}[t]{0.245\textwidth}
        \centering
        \includegraphics[width=\linewidth]{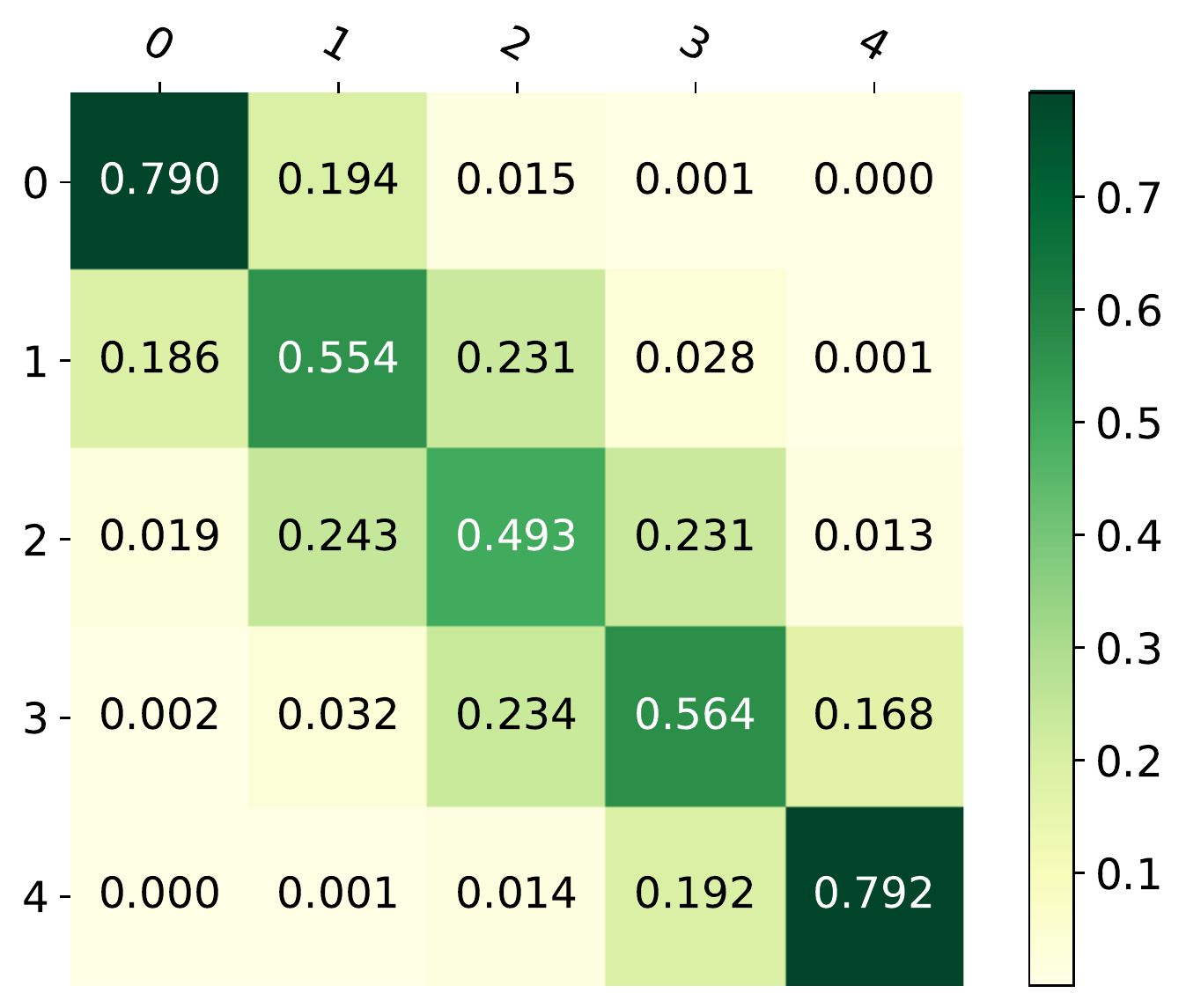}
        \caption{\mc}
    \end{subfigure}
    \begin{subfigure}[t]{0.245\textwidth}
        \centering
        \includegraphics[width=\linewidth]{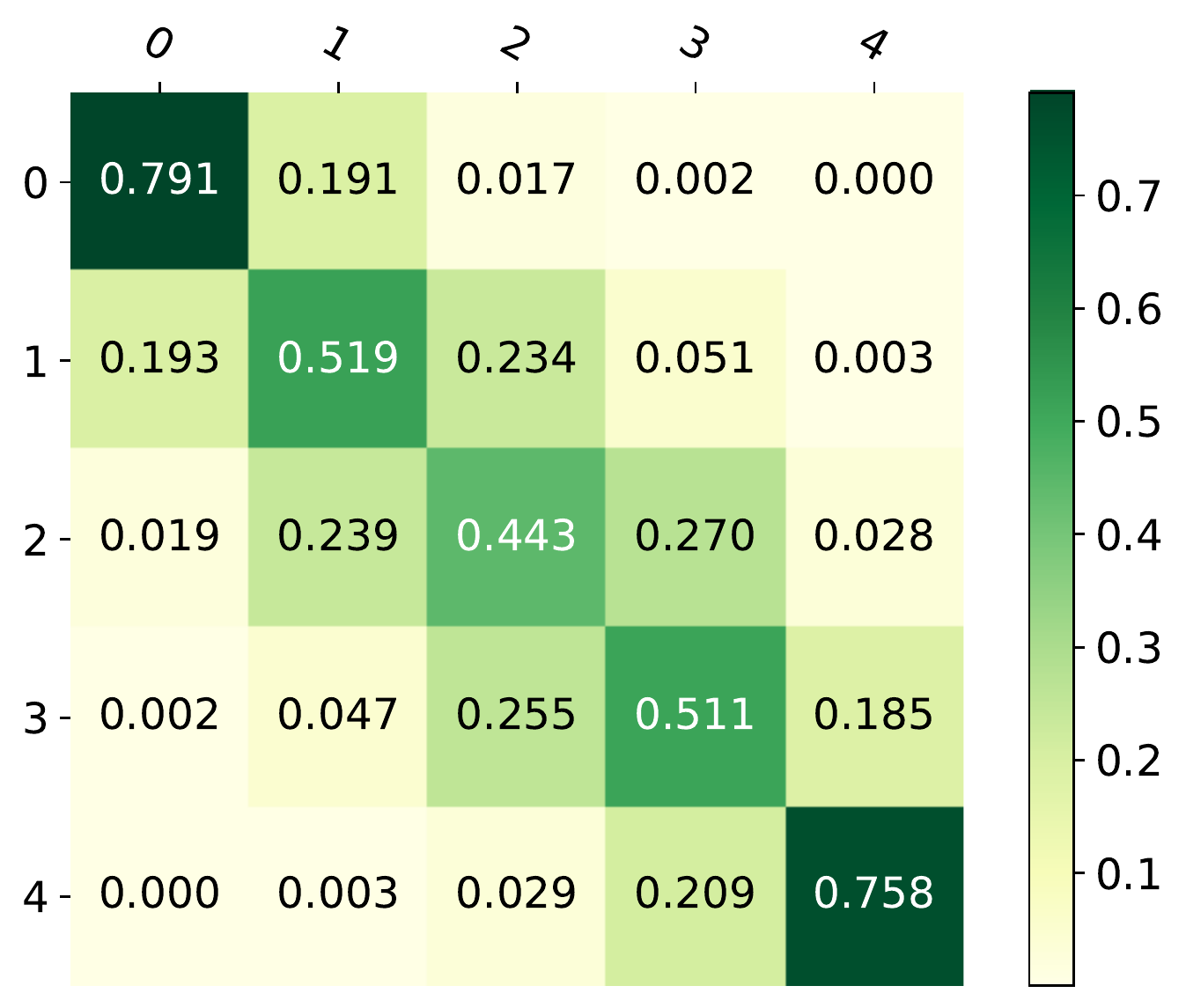}
        \caption{\mr}
    \end{subfigure}
    \caption{Confusion matrix of classification tasks for DPU estimation with dropout rate 0.1.
    The rows and columns represent labels and predictions respectively. }
    \label{fig:dropout_results_confmat}
\end{figure*}

\subsection{Target Task Performance}
\label{app:target_task_perf}
Dropout usually has quality impact. It may reduces the effective number of parameters that a model can use to produce a prediction.
Therefore, we first summarize the inference quality on the target task of the dropped out model for different dropout rates for the different datasets in Table~\ref{tab:original_perform}.
Due to space limitation, the table only includes results for dropout rate 0.1, 0.3, and 0.5.
Prediction quality is evaluated by accuracy for 
\mc\ and \emnist, by AUC for \criteo, and by mean squared error (MSE) for  \mr. Table~\ref{tab:original_perform} shows averaged results across 20 independently trained models and 100 independent dropout inferences, if dropout applies, for robustness purpose.
As expected, quality degrades with larger dropout rates but it stays about the same if we do not apply dropout at inference. 
Inference without dropout's quality is also comparable
to inference with an ensemble of size 100.

\subsection{Uncertainty Estimation}
\label{sec:dropout_estimation_performance}

Here we demonstrate the DPUs for the different datasets, and then show the performance of uncertainty estimation task models.  
We study both configurations under dropout rates
$r$ = \{0.1, 0.2, 0.3, 0.4, 0.5\}. 

\vspace{0.15cm}
{\bf DPU distribution} --- 
Figure~\ref{fig:uncer_distribution} shows histograms of DPUs for the four different tasks with dropout rate $r = 0.1$.
Table~\ref{tab:uncer_range} supplements Figure~\ref{fig:uncer_distribution} and shows the means and standard deviations of DPUs for different target tasks
over $D_{train}^{\prime}$ of randomly-initialized and randomly-shuffled ensembles and dropout models.
The distributions of DPUs appear quite different for the different datasets.  In particular, \emnist\ exhibits a rather long-tailed distribution, with most examples having relatively low prediction uncertainties and remaining examples having very large prediction uncertainties. Different digits behave differently in \emnist, as shown in Figure~\ref{fig:emnist_digitwise}.
Digit 0 is usually recognized correctly, while the digit 4 can be confused with 9. Increasing dropout rate for 0 does not change the distribution of its PU significantly. The PU distribution for 4, on the other hand, exhibits very noticeable difference when dropout rate is increased.

\vspace{0.15cm}
{\bf DPU Estimation} --- 
Tables~\ref{tab:dropout_results_regression} shows $R^2$ correlations between the predictions of the uncertainty estimation task and the true dropout PU labels.  Table~\ref{tab:dropout_results_classification} shows classification accuracy with 5 buckets for the two configurations in Figure~\ref{fig:configurations}.
Figure~\ref{fig:dropout_results_confmat} shows the confusion matrices for 
the classification task for Configuration 1 with dropout rate 0.1 for the different datasets.  Results, qualitatively similar to Figure~\ref{fig:dropout_results_confmat}, are obtained with other dropout rates. 

The results summarized in these tables and figures demonstrate that the uncertainty estimation task is able to infer the actual PU scores rather accurately in almost all cases.  We observe $R^2$ scores 0.5 or higher in most cases with regression and relatively high accuracy with classification, with the exception of \emnist. 
As we observe in Figure~\ref{fig:uncer_distribution}, DPU for \emnist\ is usually rather low.  However, high DPU examples, while sparse, can have very large DPU.  The extremely skewed DPU distribution has significant effects on the overall observed average metrics, degrading the quality of the prediction uncertainty estimation task. Also, the choice of dropout rate and the actual digit can significantly affect the actual DPU as shown in
Figure~\ref{fig:emnist_digitwise}.

Figure~\ref{fig:dropout_results_confmat} shows that the uncertainty estimation model is able to predict the top and bottom PU buckets even more accurately than other buckets,
with accuracy close to 0.8 for \movielens\ 
and even close to 0.9 for \criteo.  This implies that if one wants to classify only extreme cases, i.e., whether the model predicts very consistently or inconsistently on an example, using the low-complexity uncertainty estimation task is almost as reliable as using the expensive multiple dropout-inference method.  Given the nature of the \emnist\ dataset, the prediction accuracy of the uncertainty estimation task for the most certain bucket is rather high (around 0.8).  Because the most uncertain bucket has examples with very skewed DPU similar to the overall DPU distribution, the accuracy is therefore limited for that bucket.

\begin{figure*}
    \centering
    \begin{subfigure}[t]{0.245\textwidth}
        \centering
        \includegraphics[width=\textwidth]{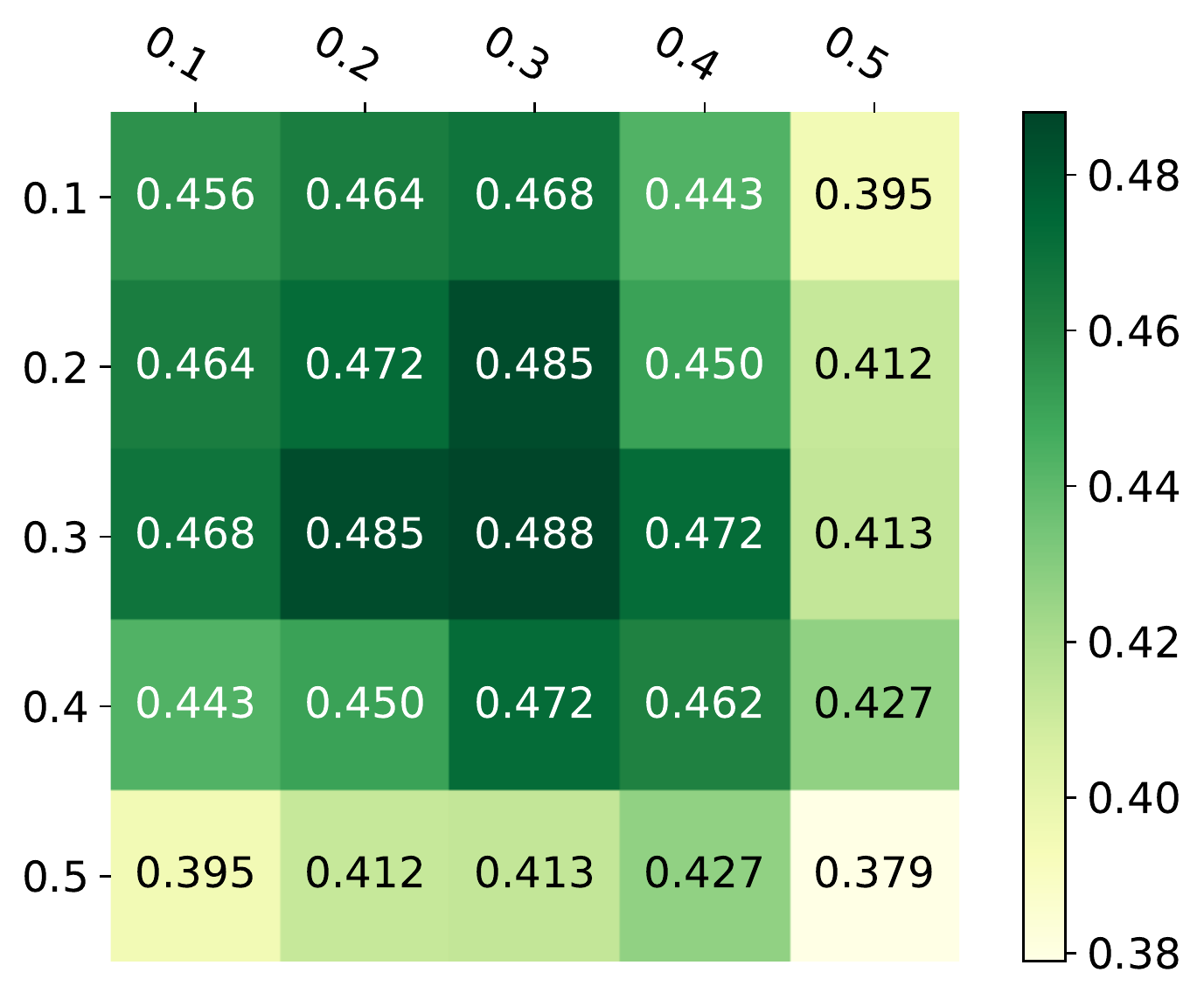}
        \caption{\emnist}
    \end{subfigure}
    \begin{subfigure}[t]{0.245\textwidth}
        \centering
        \includegraphics[width=\textwidth]{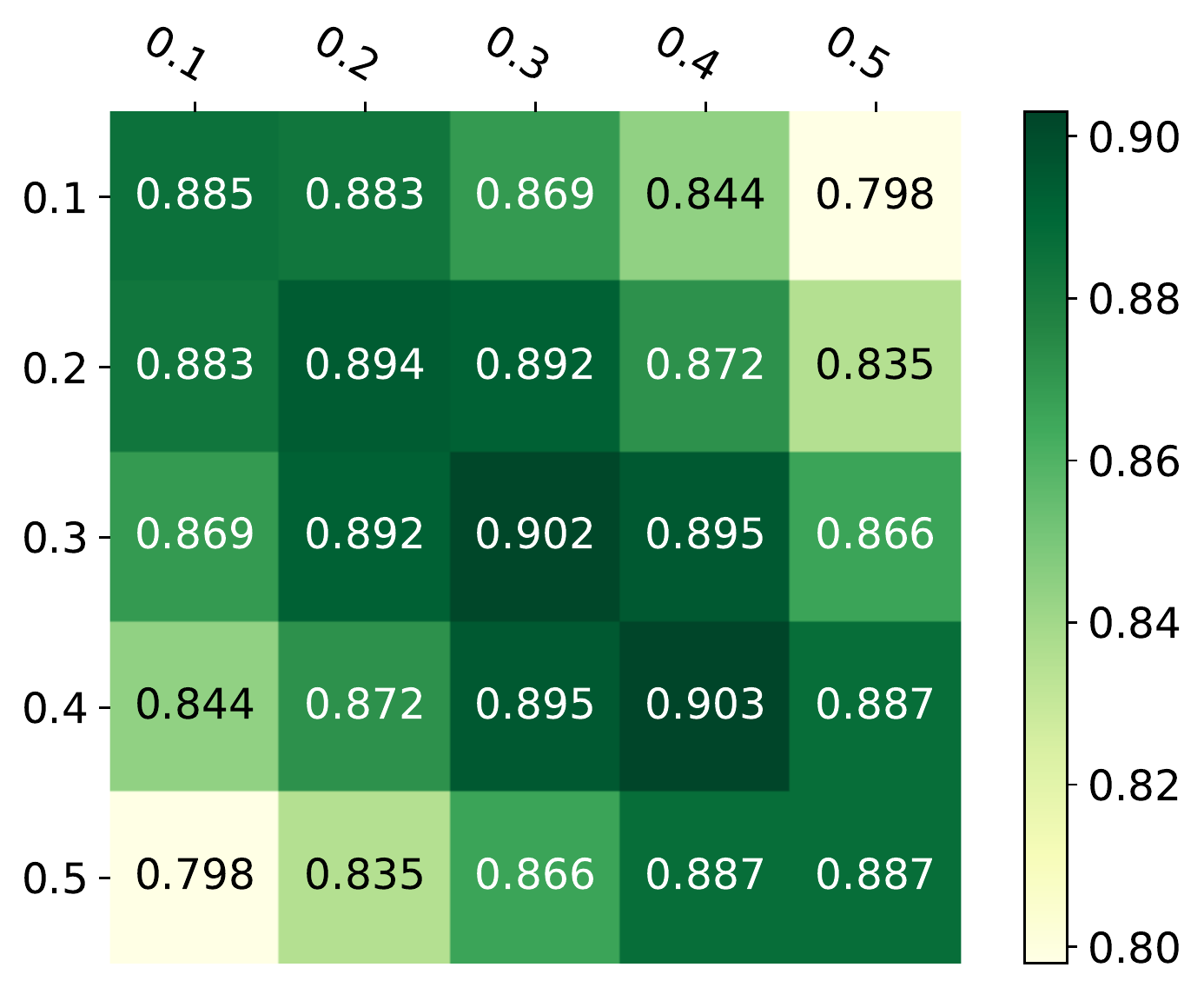}
        \caption{\criteo}
    \end{subfigure}
    \begin{subfigure}[t]{0.245\textwidth}
        \centering
        \includegraphics[width=\textwidth]{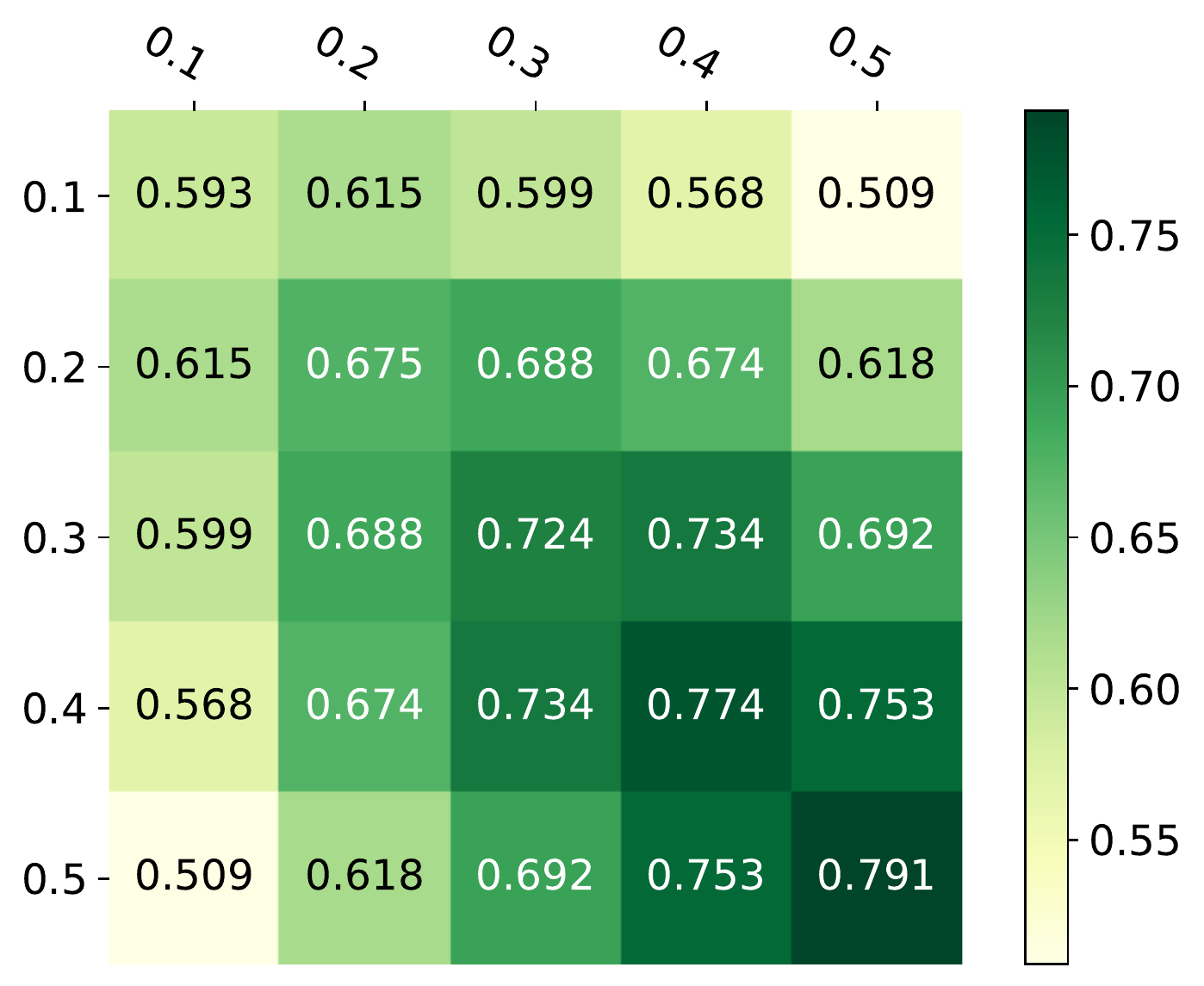}
        \caption{\mc}
    \end{subfigure}
    \begin{subfigure}[t]{0.245\textwidth}
        \centering
        \includegraphics[width=\textwidth]{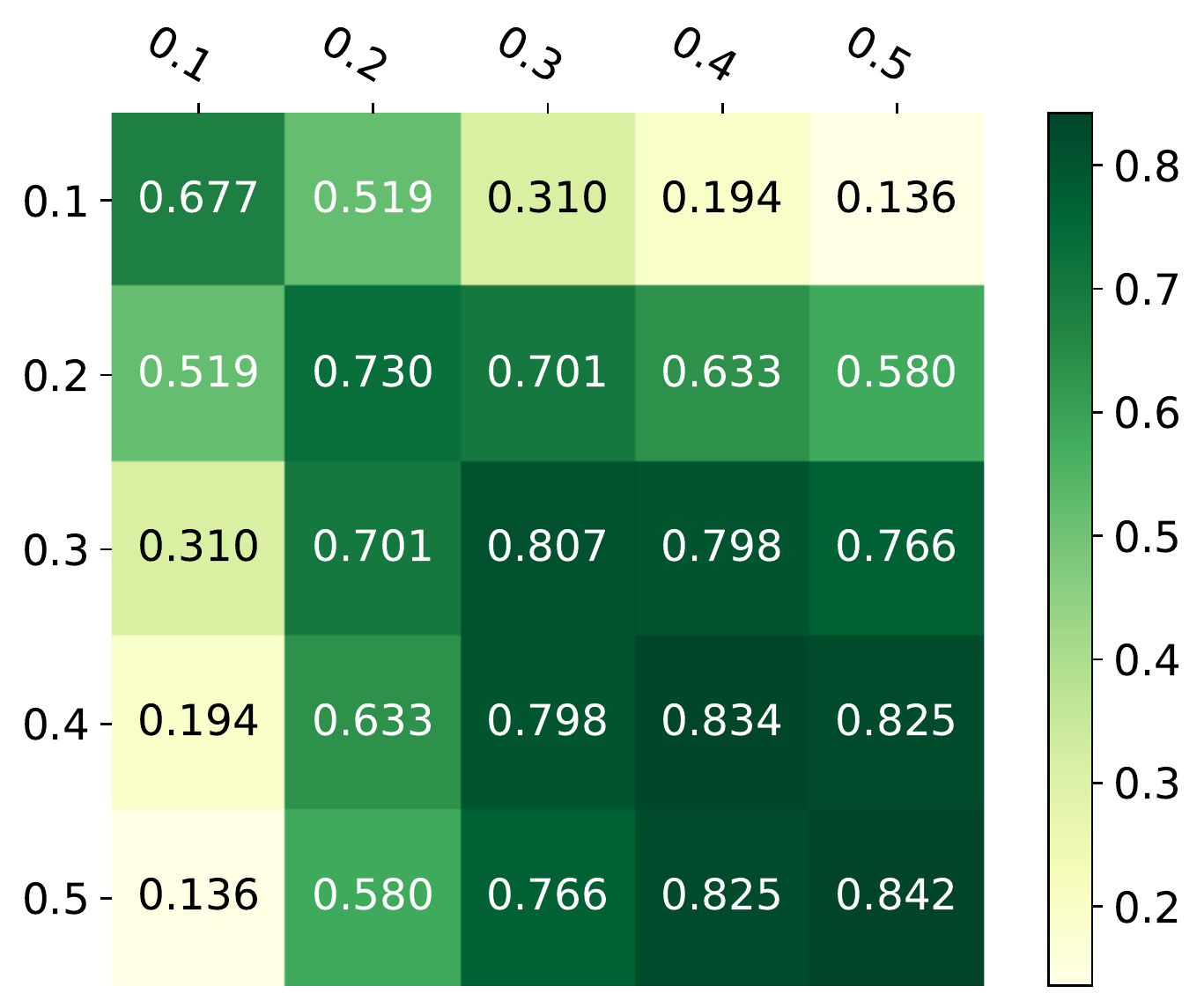}
        \caption{\mr}
    \end{subfigure}
    \caption{DPU correlation of different dropout rates. The DPU was collected by running dropout inference 100 times.}
    \label{fig:dropout}
\end{figure*}

\subsection{Dropout Prediction Uncertainty Sensitivity}
It is useful to know  whether DPUs from different instantiations of a dropout model  correlate well with each other. 
In this section, we measure DPU correlations between two independently trained target task models when: a) they use the same dropout rate, b) they use different dropout rates. These measurements give some indication of the sensitivity of the DPU to model random variations and dropout rate choices.

Figure~\ref{fig:dropout} reports $R^2$ correlations between DPUs obtained with different dropout rates averaged over 20 independent runs with different initialization, each with 100 independent dropout inferences, for the four different tasks.  Correlations for the same dropout rate are measured with two retrains of a model. As expected, DPU correlations are higher if dropout rates are the same than they're different. 
Larger dropout rate differences reduce the correlations more. Also, we can see that the dropout rate for the highest DPU correlation is different for different datasets.
This suggests that the best dropout rate may be dataset dependent. Our focus, in this paper, is on a cost-efficient method to estimate DPUs, and not on suggesting the best dropout rate for an application. 

Furthermore, we look at the effect of the number of dropout inferences on DPU correlations.
Table~\ref{tab:dropout_inference_times} shows the correlation of DPUs between two independent dropout models on \mr\ where each model uses $N \in \{10, 50, 100\}$ dropout inferences for different dropout rates. As expected, the prediction uncertainty correlation tends to 
increase with increased number of inferences as larger $N$ captures more model variations. 

\begin{table}[th]
    \centering
    \begin{tabular}{c|ccccc}
        \hline
        \multirow{2}{*}{Number of Inferences} & \multicolumn{5}{c}{Dropout Rate} \\
        \cline{2-6}
        & 0.1 & 0.2 & 0.3 & 0.4 & 0.5 \\
        \hline
        10 & 0.373 & 0.450 & 0.552 & 0.600 & 0.620 \\
        50 & 0.617 & 0.684 & 0.769 & 0.801 & 0.810 \\
        100 & 0.677 & 0.730 & 0.807 & 0.834 & 0.842 \\
        \hline
    \end{tabular}
    \caption{DPU correlation on \mr\ 
    by running inference 10, 50, and 100 times for different dropout rates.}
    \label{tab:dropout_inference_times}
\end{table}

\begin{figure*}[th]
    \centering
    \begin{subfigure}[t]{0.245\textwidth}
        \centering
        \includegraphics[width=\linewidth]{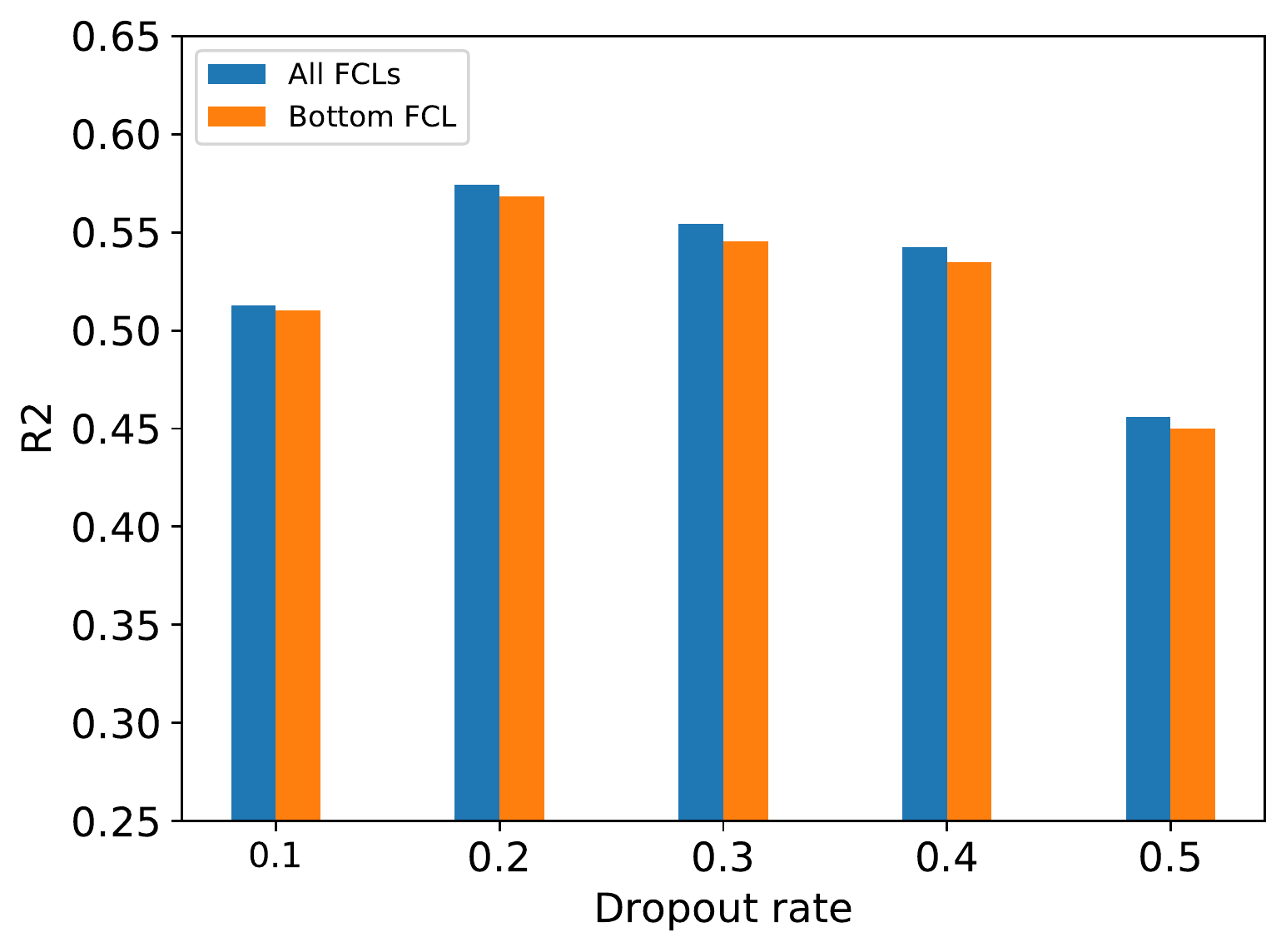}
        \caption{\emnist}
    \end{subfigure}
    \begin{subfigure}[t]{0.245\textwidth}
        \centering
        \includegraphics[width=\linewidth]{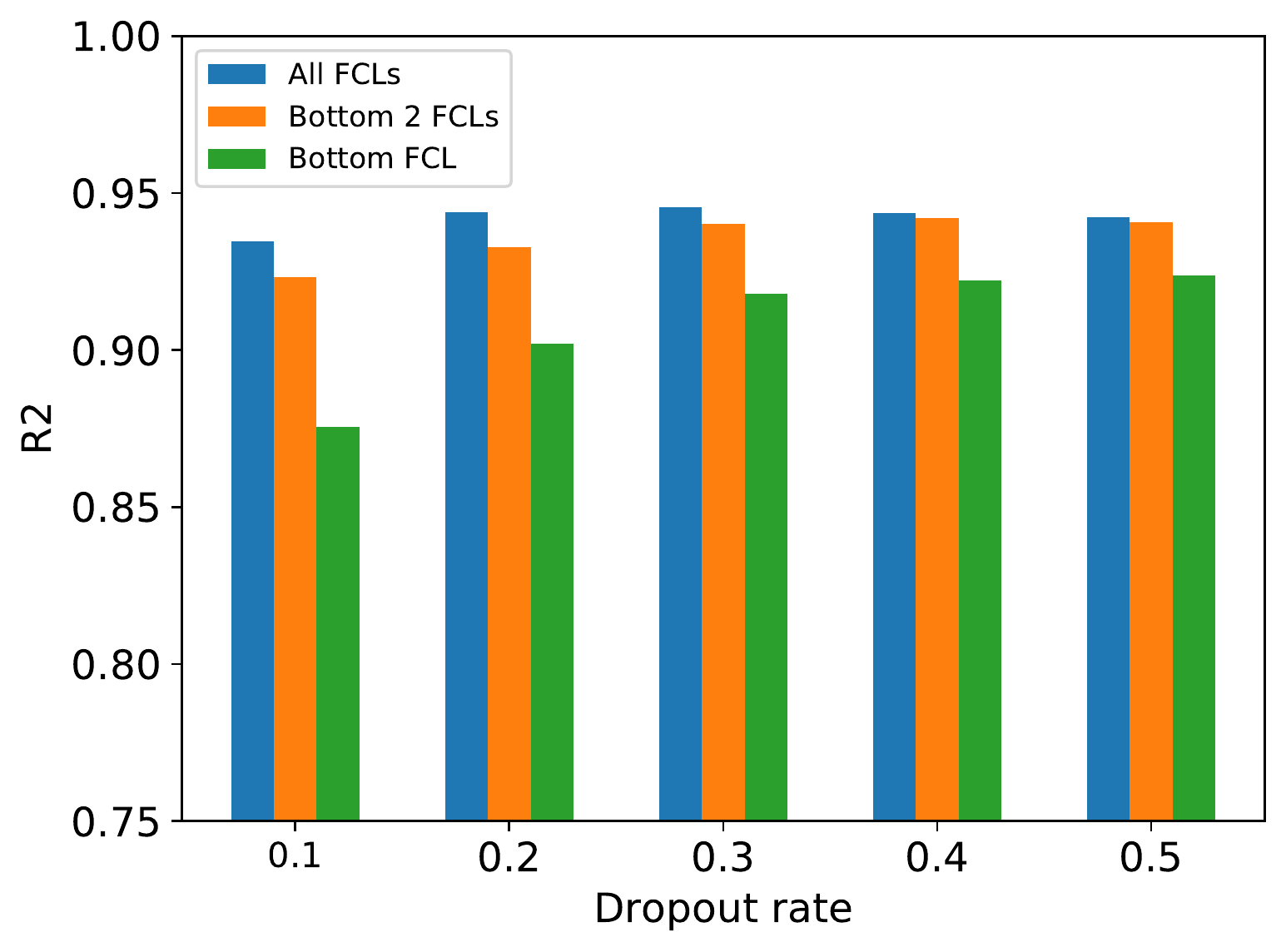}
        \caption{\criteo}
    \end{subfigure}
    \begin{subfigure}[t]{0.245\textwidth}
        \centering
        \includegraphics[width=\linewidth]{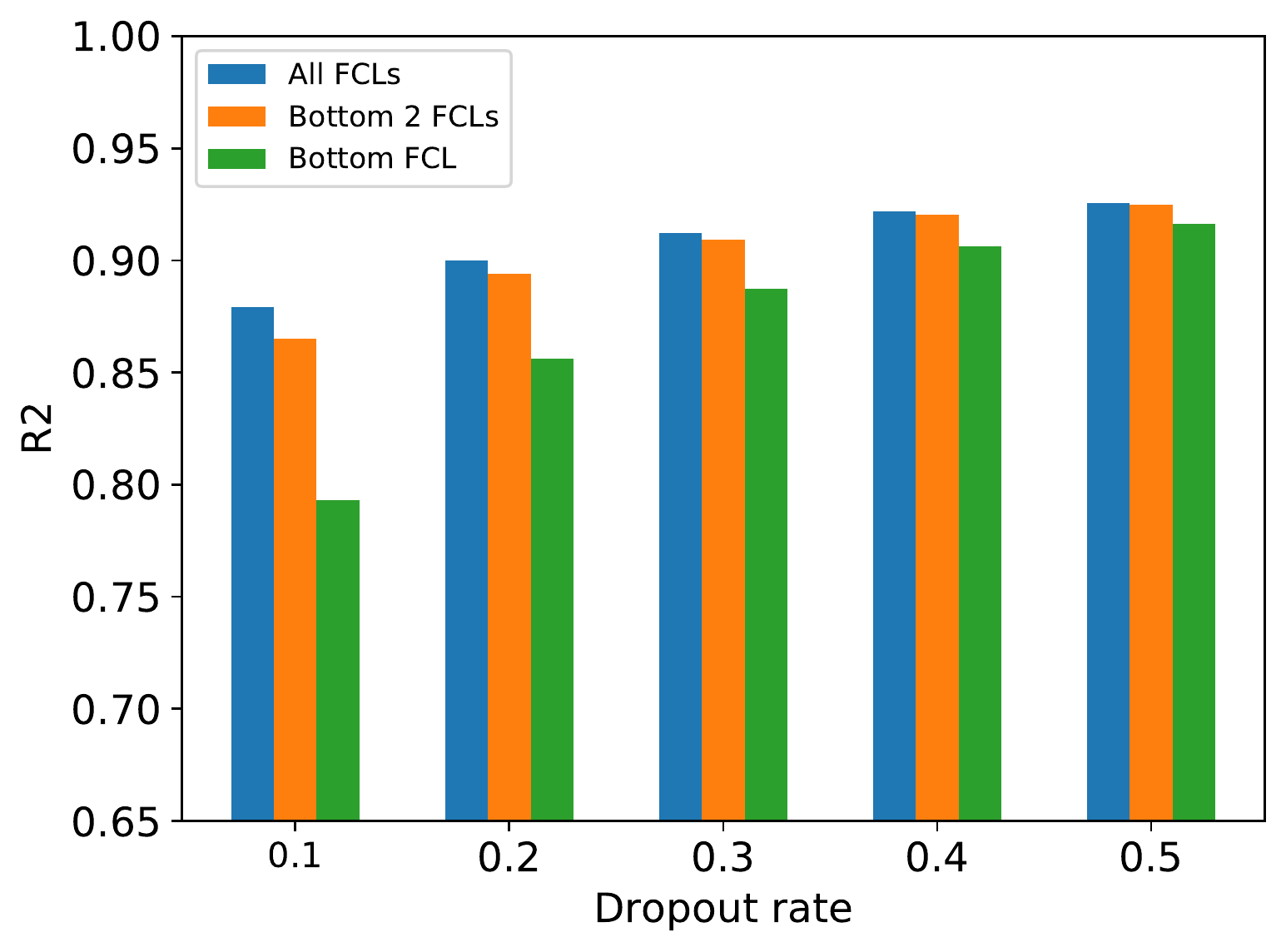}
        \caption{\mc}
    \end{subfigure}
    \begin{subfigure}[t]{0.245\textwidth}
        \centering
        \includegraphics[width=\linewidth]{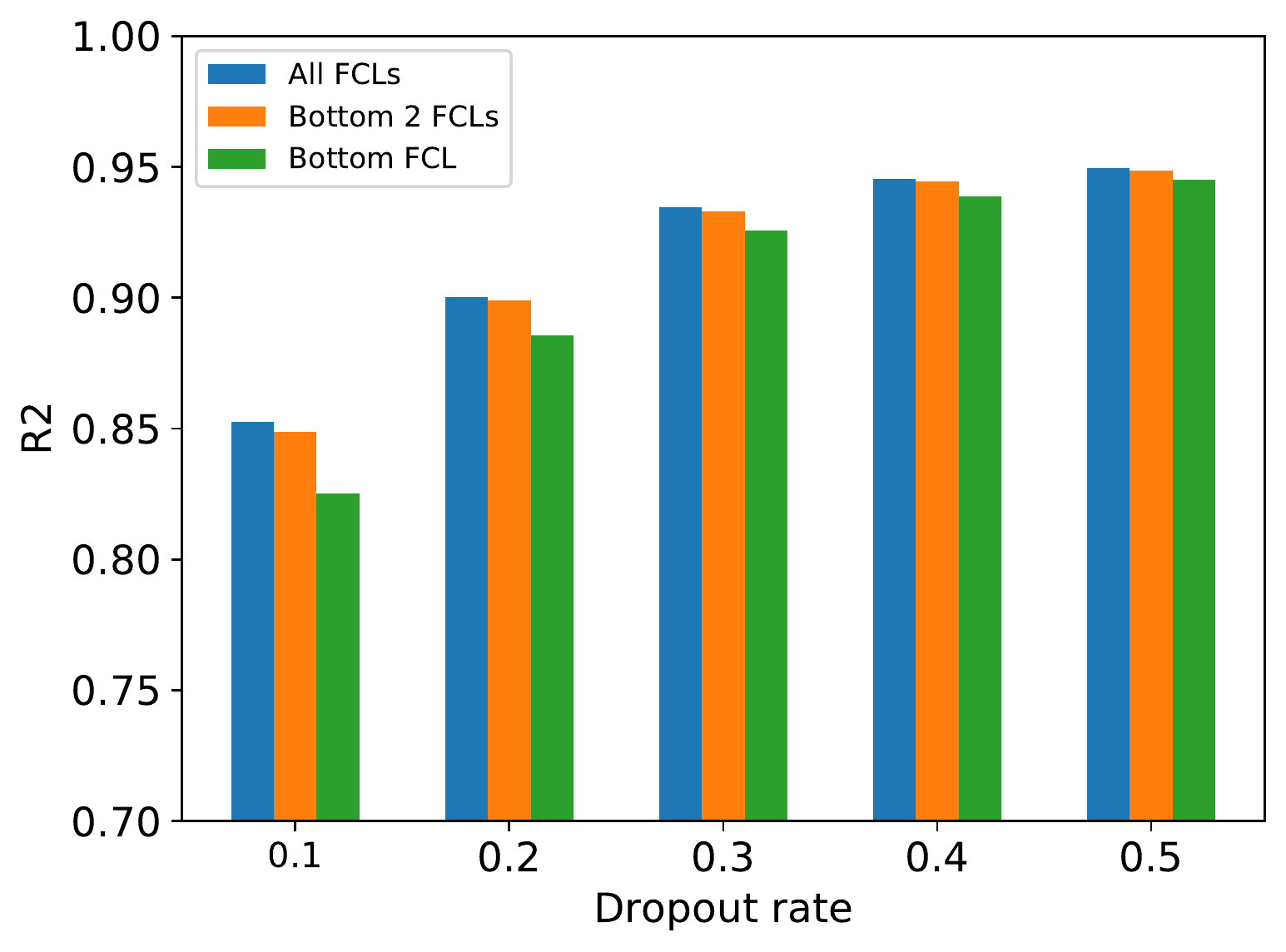}
        \caption{\mr}
    \end{subfigure}
    \caption{DPU estimation performance using activations from different subsets of FCLs.  The "All FCLs" bar is for inputs that include activations for all FCLs of the target model.
    For the "Bottom 2 FCLs" bar, inputs are from the two hidden layers closest to the input of the target task.  For the "Bottom FCL" bar, inputs are taken only from the fully-connected input layer closest to the input of the target task.}
    \label{fig:dropout_prune}
\end{figure*}

\subsection{Layer Selection}
\label{sec:pruning}
\begin{table}[t]
    \centering
    \begin{tabular}{l|c|c|c}
        \hline
        Target Tasks & All FCLs & Bottom 2 FCLs & Bottom FCL \\
        \hline
        \emnist & 45,850 & - & 29,050 (-36.6\%) \\
        \criteo & 85,050 & 75,050 (-11.8\%) & 55,050 (-35.3\%) \\
        \mc     & 85,050 & 75,050 (-11.8\%) & 55,050 (-35.3\%) \\
        \mr     & 85,050 & 75,050 (-11.8\%) & 55,050 (-35.3\%) \\
        \hline
    \end{tabular}
    \caption{Parameter totals for uncertainty estimation model with inputs from different combinations of FCLs.}
    \label{tab:flops}
\end{table}

The empirical results shown in Section~\ref{sec:dropout_estimation_performance} 
use activation strengths from all FCLs of the target task as inputs to the uncertainty estimation model. Here we show that it is sufficient to use only a subset of the layers to further reduce the cost of estimating the DPU scores.
Results are shown for Configuration 1, and we observe similar behaviors for Configuration 2.

Figure~\ref{fig:dropout_prune} shows $R^2$ correlations between the DPU label and its estimate from the uncertainty estimation model for the four target tasks, with dropout rate 0.1. Results are shown for three sets of inputs: a) activations in all fully-connected hidden layers, b) activations from the bottom two fully-connected hidden layer (where the bottom layer is the one closest to the inputs), and c) activations only from the bottom hidden layer.  Using the bottom two layers exhibits almost no degradation from using all layers for all datasets. Using a single layer also achieves rather favorable results, with a slight degradation relative to all layers.
Similar results are obtained for other dropout rates.

Table~\ref{tab:flops} shows the reduction in the number of parameters by using only the bottom two or one layer relative to using all FCL layers.
For example, for \emnist, using all FCL layers, ignoring the bias unit, we use
$2*204*100 + 100*50 + 50*1 = 45850$ hidden parameters for the uncertainty estimation task, with 2*204 inputs (204 neurons of both binary and value features), and a network of sizes [100, 50, 1].
Using only the activations of the bottom fully-connected hidden layer, number of parameters of the uncertainty estimation model can be reduced by over 35\%, which corresponds to a 35\% reduction of computation. 

Further complexity reductions are possible by using inputs from layers with fewer neurons (closer to the output of the target task).  Using these inputs does show slight degradation in $R^2$, but not substantial.  
For example, we experimented with using neuron activation inputs from any single FCL on \criteo.
The $R^2$ correlations between the estimated DPU and the DPU labels were 0.882, 0.875, or 0.837 for the 250-neuron layer, 100-neuron layer, or 50-neuron layer,
respectively.

Overall, our results demonstrate that by leveraging neuron activation strengths, 
our uncertainty estimation task, as a cost-efficient auxiliary model, is able to reduce the need for multiple inferences for DPU estimation. 
Moreover, our DPU estimates can be rather reliably obtained using activations from even a single hidden neuron layer from the target task.  
A designer can tune the complexity to balance accuracy on the DPU estimation to fit the needs of their application.

\section{Discussion: Dropout Vs. Ensemble}
\label{sec:discussion}
Deep neural networks have complex loss landscapes with multiple local optima.
Convergence to different optima is one cause of prediction uncertainties that are described in Section~\ref{sec:intro}.
However, uncertainties can be observed even if
a training model is on track to converge to the same optimum but may have not reached it yet for some reason, e.g, early stopping, stuck at a local minima, oscillating around the optimum, etc.
Prediction uncertainties can be
due to data uncertainties, model misspecification, algorithm/optimizer suboptimality, training data sampling strategies, and other factors.  A model can approach the same optimum or different optima from different directions, depending on initialization, order of training examples, schedule of its updates, random rounding errors, distributed training and parallelization, and even different software and hardware effects on the same algorithm. For example, some properties of a data center can affect order of operations which can result in one order in one data center and another in another.

The dropout prediction uncertainty for a given example is obtained from a collection of predictions. Each prediction is generated by randomly dropping neurons from the trained model at a given rate. Each time, the remaining neurons effectively constitute a slightly different model. 
Dropout in training ensures that the inference of a dropped out model is robust and is able to produce reliable predictions despite the dropout.
Higher dropout rates introduce larger diversities.
Table {\ref{tab:uncer_range}} in Section~\ref{app:target_task_perf}
demonstrates that both prediction uncertainty mean and variance increase when dropout rate increases.

Following the above dropout description, it's natural to view dropout as a way of constructing an ensemble (see also \cite{srivastava2014dropout, hara2016analysis, gal2015bayesian}). The members of a dropout ensemble differ from but also correlate with each other because the dropout model parameters are updated in tandem during training. Therefore, the prediction uncertainties from dropout ensemble members can be different from those from independently trained ensemble members with other differences, like different random initializations. Different diversification of ensemble members produces different prediction uncertainties~\cite{chen2021beyond,snapp2021synthesizing}. Some open questions remain to be answered: 1) what uncertainty factors contribute significantly to dropout prediction uncertainties; 2) how dropout prediction uncertainties correspond to prediction uncertainties introduced by other ensemble diversification schemes.
\section{Conclusion and Future Work}
\label{sec:conclusion}

We proposed and studied low-cost low-complexity prediction uncertainty estimators that predict prediction uncertainties obtained by dropouts in a deep network.
Dropout prediction uncertainty measurements can be obtained by multiple inferences of a model on a single example, where each inference selects a different set of neurons to be dropped out.  
Our proposed estimator trains on such dropout prediction uncertainty measurements and requires only a single forward pass at inference time. 
We demonstrated on three benchmark datasets that this estimator gives reliable estimates of the dropout prediction uncertainties without the need of multiple inferences.  Empirical results showed a high correlation of the estimated scores to DPUs in regression mode, and high accuracy in classification mode, in which DPUs were clustered into buckets of different uncertainty levels.  These results were shown to be attainable even if only neurons from subsets of the network layers of the target task were used as inputs to the uncertainty estimation task, and even when neuron activations from only a single layer of the target task were used.
Our result is a proof of concept that a practical low-cost and low-complexity methodology can be used to output prediction uncertainties for deep networks.

Future work could focus on the composition of the prediction uncertainty sources, on how such compositions are reflected in different prediction uncertainty measures, and on how estimation of such measures can be simplified.  Understanding the differences between different measures can give rise to the ability to isolate each of the sources of prediction uncertainty, perhaps to a point that will enable mitigation of such uncertainties if necessary.  It can also lead to identifying the sources in different applications (sources that may be different or may have different levels of influence on the overall uncertainties in different applications).
It can also lead to a better understanding of which types of estimators, like the one we proposed in this paper, are effective for which sources of prediction uncertainty.  Specifically, it would be beneficial to develop an understanding of the differences between ensemble-based uncertainty measurements and dropout uncertainty measurements, as well as the differences within different techniques that are used for ensembles, and within different dropout strategies.
Finally, generalizing our approach to problems, such as reinforcement and curriculum learning, is another interesting direction.

\bibliographystyle{ACM-Reference-Format}
\bibliography{8_reference}


\begin{thebibliography}{54}


\ifx \showCODEN    \undefined \def \showCODEN     #1{\unskip}     \fi
\ifx \showDOI      \undefined \def \showDOI       #1{#1}\fi
\ifx \showISBNx    \undefined \def \showISBNx     #1{\unskip}     \fi
\ifx \showISBNxiii \undefined \def \showISBNxiii  #1{\unskip}     \fi
\ifx \showISSN     \undefined \def \showISSN      #1{\unskip}     \fi
\ifx \showLCCN     \undefined \def \showLCCN      #1{\unskip}     \fi
\ifx \shownote     \undefined \def \shownote      #1{#1}          \fi
\ifx \showarticletitle \undefined \def \showarticletitle #1{#1}   \fi
\ifx \showURL      \undefined \def \showURL       {\relax}        \fi
\providecommand\bibfield[2]{#2}
\providecommand\bibinfo[2]{#2}
\providecommand\natexlab[1]{#1}
\providecommand\showeprint[2][]{arXiv:#2}

\bibitem[\protect\citeauthoryear{Achrack, Kellerman, and Barzilay}{Achrack
  et~al\mbox{.}}{2020}]%
        {achrack2020multi}
\bibfield{author}{\bibinfo{person}{Omer Achrack}, \bibinfo{person}{Raizy
  Kellerman}, {and} \bibinfo{person}{Ouriel Barzilay}.}
  \bibinfo{year}{2020}\natexlab{}.
\newblock \showarticletitle{Multi-Loss Sub-Ensembles for Accurate
  Classification with Uncertainty Estimation}.
\newblock \bibinfo{journal}{\emph{arXiv preprint arXiv:2010.01917}}
  (\bibinfo{year}{2020}).
\newblock


\bibitem[\protect\citeauthoryear{Allen-Zhu and Li}{Allen-Zhu and Li}{2021}]%
        {allenzhu21towards}
\bibfield{author}{\bibinfo{person}{Zeyuan Allen-Zhu} {and}
  \bibinfo{person}{Yuanzhi Li}.} \bibinfo{year}{2021}\natexlab{}.
\newblock \showarticletitle{Towards Understanding Ensemble, Knowledge
  Distillation and Self-Distillation in Deep Learning}.
\newblock \bibinfo{journal}{\emph{arXiv preprint arXiv:2012.09816}}
  (\bibinfo{year}{2021}).
\newblock


\bibitem[\protect\citeauthoryear{Anil, Pereyra, Passos, Ormandi, Dahl, and
  Hinton}{Anil et~al\mbox{.}}{2018}]%
        {anil18large}
\bibfield{author}{\bibinfo{person}{Rohan Anil}, \bibinfo{person}{Gabriel
  Pereyra}, \bibinfo{person}{Alexandre Passos}, \bibinfo{person}{Robert
  Ormandi}, \bibinfo{person}{George~E Dahl}, {and} \bibinfo{person}{Geoffrey~E
  Hinton}.} \bibinfo{year}{2018}\natexlab{}.
\newblock \showarticletitle{Large scale distributed neural network training
  through online distillation}.
\newblock \bibinfo{journal}{\emph{arXiv preprint arXiv:1804.03235}}
  (\bibinfo{year}{2018}).
\newblock


\bibitem[\protect\citeauthoryear{Atencia, Stoean, and Joya}{Atencia
  et~al\mbox{.}}{2020}]%
        {atencia2020uncertainty}
\bibfield{author}{\bibinfo{person}{Miguel Atencia}, \bibinfo{person}{Ruxandra
  Stoean}, {and} \bibinfo{person}{Gonzalo Joya}.}
  \bibinfo{year}{2020}\natexlab{}.
\newblock \showarticletitle{Uncertainty Quantification through Dropout in Time
  Series Prediction by Echo State Networks}.
\newblock \bibinfo{journal}{\emph{Mathematics}} \bibinfo{volume}{8},
  \bibinfo{number}{8} (\bibinfo{year}{2020}), \bibinfo{pages}{1374}.
\newblock


\bibitem[\protect\citeauthoryear{Blundell, Cornebise, Kavukcuoglu, and
  Wierstra}{Blundell et~al\mbox{.}}{2015}]%
        {blundell2015weight}
\bibfield{author}{\bibinfo{person}{Charles Blundell}, \bibinfo{person}{Julien
  Cornebise}, \bibinfo{person}{Koray Kavukcuoglu}, {and} \bibinfo{person}{Daan
  Wierstra}.} \bibinfo{year}{2015}\natexlab{}.
\newblock \showarticletitle{Weight uncertainty in neural networks}.
\newblock \bibinfo{journal}{\emph{arXiv preprint arXiv:1505.05424}}
  (\bibinfo{year}{2015}).
\newblock


\bibitem[\protect\citeauthoryear{Chen, Wang, Lin, Cheng, Hong, Chi, and
  Cui}{Chen et~al\mbox{.}}{2021}]%
        {chen2021beyond}
\bibfield{author}{\bibinfo{person}{Zhe Chen}, \bibinfo{person}{Yuyan Wang},
  \bibinfo{person}{Dong Lin}, \bibinfo{person}{Derek~Zhiyuan Cheng},
  \bibinfo{person}{Lichan Hong}, \bibinfo{person}{Ed~H Chi}, {and}
  \bibinfo{person}{Claire Cui}.} \bibinfo{year}{2021}\natexlab{}.
\newblock \showarticletitle{Beyond Point Estimate: Inferring Ensemble
  Prediction Variation from Neuron Activation Strength in Recommender Systems}.
  In \bibinfo{booktitle}{\emph{Proceedings of the 14th ACM International
  Conference on Web Search and Data Mining}}. \bibinfo{pages}{76--84}.
\newblock


\bibitem[\protect\citeauthoryear{Cohen, Afshar, Tapson, and Van~Schaik}{Cohen
  et~al\mbox{.}}{2017}]%
        {cohen2017emnist}
\bibfield{author}{\bibinfo{person}{Gregory Cohen}, \bibinfo{person}{Saeed
  Afshar}, \bibinfo{person}{Jonathan Tapson}, {and} \bibinfo{person}{Andre
  Van~Schaik}.} \bibinfo{year}{2017}\natexlab{}.
\newblock \showarticletitle{EMNIST: Extending MNIST to handwritten letters}. In
  \bibinfo{booktitle}{\emph{2017 International Joint Conference on Neural
  Networks (IJCNN)}}. IEEE, \bibinfo{pages}{2921--2926}.
\newblock


\bibitem[\protect\citeauthoryear{Covington, Adams, and Sargin}{Covington
  et~al\mbox{.}}{2016}]%
        {covington2016deep}
\bibfield{author}{\bibinfo{person}{Paul Covington}, \bibinfo{person}{Jay
  Adams}, {and} \bibinfo{person}{Emre Sargin}.}
  \bibinfo{year}{2016}\natexlab{}.
\newblock \showarticletitle{Deep neural networks for youtube recommendations}.
  In \bibinfo{booktitle}{\emph{Proceedings of the 10th ACM conference on
  recommender systems}}. \bibinfo{pages}{191--198}.
\newblock


\bibitem[\protect\citeauthoryear{D'Amour, Heller, Moldovan, Adlam, Alipanahi,
  Beutel, Chen, Deaton, Eisenstein, Hoffman, et~al\mbox{.}}{D'Amour
  et~al\mbox{.}}{2020}]%
        {d2020underspecification}
\bibfield{author}{\bibinfo{person}{Alexander D'Amour},
  \bibinfo{person}{Katherine Heller}, \bibinfo{person}{Dan Moldovan},
  \bibinfo{person}{Ben Adlam}, \bibinfo{person}{Babak Alipanahi},
  \bibinfo{person}{Alex Beutel}, \bibinfo{person}{Christina Chen},
  \bibinfo{person}{Jonathan Deaton}, \bibinfo{person}{Jacob Eisenstein},
  \bibinfo{person}{Matthew~D Hoffman}, {et~al\mbox{.}}}
  \bibinfo{year}{2020}\natexlab{}.
\newblock \showarticletitle{Underspecification presents challenges for
  credibility in modern machine learning}.
\newblock \bibinfo{journal}{\emph{arXiv preprint arXiv:2011.03395}}
  (\bibinfo{year}{2020}).
\newblock


\bibitem[\protect\citeauthoryear{Eaton-Rosen, Bragman, Bisdas, Ourselin, and
  Cardoso}{Eaton-Rosen et~al\mbox{.}}{2018}]%
        {eaton2018towards}
\bibfield{author}{\bibinfo{person}{Zach Eaton-Rosen}, \bibinfo{person}{Felix
  Bragman}, \bibinfo{person}{Sotirios Bisdas}, \bibinfo{person}{S{\'e}bastien
  Ourselin}, {and} \bibinfo{person}{M~Jorge Cardoso}.}
  \bibinfo{year}{2018}\natexlab{}.
\newblock \showarticletitle{Towards safe deep learning: accurately quantifying
  biomarker uncertainty in neural network predictions}. In
  \bibinfo{booktitle}{\emph{International Conference on Medical Image Computing
  and Computer-Assisted Intervention}}. Springer, \bibinfo{pages}{691--699}.
\newblock


\bibitem[\protect\citeauthoryear{Feng, Rosenbaum, and Dietmayer}{Feng
  et~al\mbox{.}}{2018}]%
        {feng2018towards}
\bibfield{author}{\bibinfo{person}{Di Feng}, \bibinfo{person}{Lars Rosenbaum},
  {and} \bibinfo{person}{Klaus Dietmayer}.} \bibinfo{year}{2018}\natexlab{}.
\newblock \showarticletitle{Towards safe autonomous driving: Capture
  uncertainty in the deep neural network for lidar 3d vehicle detection}. In
  \bibinfo{booktitle}{\emph{2018 21st International Conference on Intelligent
  Transportation Systems (ITSC)}}. IEEE, \bibinfo{pages}{3266--3273}.
\newblock


\bibitem[\protect\citeauthoryear{Fort, Hu, and Lakshminarayanan}{Fort
  et~al\mbox{.}}{2019}]%
        {fort2019deep}
\bibfield{author}{\bibinfo{person}{Stanislav Fort}, \bibinfo{person}{Huiyi Hu},
  {and} \bibinfo{person}{Balaji Lakshminarayanan}.}
  \bibinfo{year}{2019}\natexlab{}.
\newblock \showarticletitle{Deep ensembles: A loss landscape perspective}.
\newblock \bibinfo{journal}{\emph{arXiv preprint arXiv:1912.02757}}
  (\bibinfo{year}{2019}).
\newblock


\bibitem[\protect\citeauthoryear{Gal and Ghahramani}{Gal and
  Ghahramani}{2015}]%
        {gal2015bayesian}
\bibfield{author}{\bibinfo{person}{Yarin Gal} {and} \bibinfo{person}{Zoubin
  Ghahramani}.} \bibinfo{year}{2015}\natexlab{}.
\newblock \showarticletitle{Bayesian convolutional neural networks with
  Bernoulli approximate variational inference}.
\newblock \bibinfo{journal}{\emph{arXiv preprint arXiv:1506.02158}}
  (\bibinfo{year}{2015}).
\newblock


\bibitem[\protect\citeauthoryear{Gal and Ghahramani}{Gal and
  Ghahramani}{2016}]%
        {gal2016dropout}
\bibfield{author}{\bibinfo{person}{Yarin Gal} {and} \bibinfo{person}{Zoubin
  Ghahramani}.} \bibinfo{year}{2016}\natexlab{}.
\newblock \showarticletitle{Dropout as a bayesian approximation: Representing
  model uncertainty in deep learning}. In
  \bibinfo{booktitle}{\emph{international conference on machine learning}}.
  \bibinfo{pages}{1050--1059}.
\newblock


\bibitem[\protect\citeauthoryear{Gal, Hron, and Kendall}{Gal
  et~al\mbox{.}}{2017a}]%
        {gal2017concrete}
\bibfield{author}{\bibinfo{person}{Yarin Gal}, \bibinfo{person}{Jiri Hron},
  {and} \bibinfo{person}{Alex Kendall}.} \bibinfo{year}{2017}\natexlab{a}.
\newblock \showarticletitle{Concrete dropout}.
\newblock \bibinfo{journal}{\emph{arXiv preprint arXiv:1705.07832}}
  (\bibinfo{year}{2017}).
\newblock


\bibitem[\protect\citeauthoryear{Gal, Islam, and Ghahramani}{Gal
  et~al\mbox{.}}{2017b}]%
        {gal2017deep}
\bibfield{author}{\bibinfo{person}{Yarin Gal}, \bibinfo{person}{Riashat Islam},
  {and} \bibinfo{person}{Zoubin Ghahramani}.} \bibinfo{year}{2017}\natexlab{b}.
\newblock \showarticletitle{Deep bayesian active learning with image data}. In
  \bibinfo{booktitle}{\emph{International Conference on Machine Learning}}.
  PMLR, \bibinfo{pages}{1183--1192}.
\newblock


\bibitem[\protect\citeauthoryear{Gawlikowski, Tassi, Ali, Lee, Humt, Feng,
  Kruspe, Triebel, Jung, Roscher, et~al\mbox{.}}{Gawlikowski
  et~al\mbox{.}}{2021}]%
        {gawlikowski2021survey}
\bibfield{author}{\bibinfo{person}{Jakob Gawlikowski},
  \bibinfo{person}{Cedrique Rovile~Njieutcheu Tassi}, \bibinfo{person}{Mohsin
  Ali}, \bibinfo{person}{Jongseok Lee}, \bibinfo{person}{Matthias Humt},
  \bibinfo{person}{Jianxiang Feng}, \bibinfo{person}{Anna Kruspe},
  \bibinfo{person}{Rudolph Triebel}, \bibinfo{person}{Peter Jung},
  \bibinfo{person}{Ribana Roscher}, {et~al\mbox{.}}}
  \bibinfo{year}{2021}\natexlab{}.
\newblock \showarticletitle{A survey of uncertainty in deep neural networks}.
\newblock \bibinfo{journal}{\emph{arXiv preprint arXiv:2107.03342}}
  (\bibinfo{year}{2021}).
\newblock


\bibitem[\protect\citeauthoryear{Hara, Saitoh, and Shouno}{Hara
  et~al\mbox{.}}{2016}]%
        {hara2016analysis}
\bibfield{author}{\bibinfo{person}{Kazuyuki Hara}, \bibinfo{person}{Daisuke
  Saitoh}, {and} \bibinfo{person}{Hayaru Shouno}.}
  \bibinfo{year}{2016}\natexlab{}.
\newblock \showarticletitle{Analysis of dropout learning regarded as ensemble
  learning}. In \bibinfo{booktitle}{\emph{International Conference on
  Artificial Neural Networks}}. Springer, \bibinfo{pages}{72--79}.
\newblock


\bibitem[\protect\citeauthoryear{Harper and Konstan}{Harper and
  Konstan}{2015}]%
        {harper2015movielens}
\bibfield{author}{\bibinfo{person}{F~Maxwell Harper} {and}
  \bibinfo{person}{Joseph~A Konstan}.} \bibinfo{year}{2015}\natexlab{}.
\newblock \showarticletitle{The movielens datasets: History and context}.
\newblock \bibinfo{journal}{\emph{Acm transactions on interactive intelligent
  systems (tiis)}} \bibinfo{volume}{5}, \bibinfo{number}{4}
  (\bibinfo{year}{2015}), \bibinfo{pages}{1--19}.
\newblock


\bibitem[\protect\citeauthoryear{Havasi, Jenatton, Fort, Liu, Snoek,
  Lakshminarayanan, Dai, and Tran}{Havasi et~al\mbox{.}}{2020}]%
        {havasi2020training}
\bibfield{author}{\bibinfo{person}{Marton Havasi}, \bibinfo{person}{Rodolphe
  Jenatton}, \bibinfo{person}{Stanislav Fort}, \bibinfo{person}{Jeremiah~Zhe
  Liu}, \bibinfo{person}{Jasper Snoek}, \bibinfo{person}{Balaji
  Lakshminarayanan}, \bibinfo{person}{Andrew~M Dai}, {and}
  \bibinfo{person}{Dustin Tran}.} \bibinfo{year}{2020}\natexlab{}.
\newblock \showarticletitle{Training independent subnetworks for robust
  prediction}.
\newblock \bibinfo{journal}{\emph{arXiv preprint arXiv:2010.06610}}
  (\bibinfo{year}{2020}).
\newblock


\bibitem[\protect\citeauthoryear{He, Zhang, Ren, and Sun}{He
  et~al\mbox{.}}{2016}]%
        {he2016deep}
\bibfield{author}{\bibinfo{person}{Kaiming He}, \bibinfo{person}{Xiangyu
  Zhang}, \bibinfo{person}{Shaoqing Ren}, {and} \bibinfo{person}{Jian Sun}.}
  \bibinfo{year}{2016}\natexlab{}.
\newblock \showarticletitle{Deep residual learning for image recognition}. In
  \bibinfo{booktitle}{\emph{Proceedings of the IEEE conference on computer
  vision and pattern recognition}}. \bibinfo{pages}{770--778}.
\newblock


\bibitem[\protect\citeauthoryear{He, Liao, Zhang, Nie, Hu, and Chua}{He
  et~al\mbox{.}}{2017}]%
        {he2017neural}
\bibfield{author}{\bibinfo{person}{Xiangnan He}, \bibinfo{person}{Lizi Liao},
  \bibinfo{person}{Hanwang Zhang}, \bibinfo{person}{Liqiang Nie},
  \bibinfo{person}{Xia Hu}, {and} \bibinfo{person}{Tat-Seng Chua}.}
  \bibinfo{year}{2017}\natexlab{}.
\newblock \showarticletitle{Neural collaborative filtering}. In
  \bibinfo{booktitle}{\emph{Proceedings of the 26th international conference on
  world wide web}}. International World Wide Web Conferences Steering
  Committee, \bibinfo{pages}{173--182}.
\newblock


\bibitem[\protect\citeauthoryear{Hoffman, Blei, Wang, and Paisley}{Hoffman
  et~al\mbox{.}}{2013}]%
        {hoffman2013stochastic}
\bibfield{author}{\bibinfo{person}{Matthew~D Hoffman}, \bibinfo{person}{David~M
  Blei}, \bibinfo{person}{Chong Wang}, {and} \bibinfo{person}{John Paisley}.}
  \bibinfo{year}{2013}\natexlab{}.
\newblock \showarticletitle{Stochastic variational inference.}
\newblock \bibinfo{journal}{\emph{Journal of Machine Learning Research}}
  \bibinfo{volume}{14}, \bibinfo{number}{5} (\bibinfo{year}{2013}).
\newblock


\bibitem[\protect\citeauthoryear{Huang, Liu, Van Der~Maaten, and
  Weinberger}{Huang et~al\mbox{.}}{2017}]%
        {huang2017densely}
\bibfield{author}{\bibinfo{person}{Gao Huang}, \bibinfo{person}{Zhuang Liu},
  \bibinfo{person}{Laurens Van Der~Maaten}, {and} \bibinfo{person}{Kilian~Q
  Weinberger}.} \bibinfo{year}{2017}\natexlab{}.
\newblock \showarticletitle{Densely connected convolutional networks}. In
  \bibinfo{booktitle}{\emph{Proceedings of the IEEE conference on computer
  vision and pattern recognition}}. \bibinfo{pages}{4700--4708}.
\newblock


\bibitem[\protect\citeauthoryear{Ioffe and Szegedy}{Ioffe and Szegedy}{2015}]%
        {ioffe2015batch}
\bibfield{author}{\bibinfo{person}{Sergey Ioffe} {and}
  \bibinfo{person}{Christian Szegedy}.} \bibinfo{year}{2015}\natexlab{}.
\newblock \showarticletitle{Batch normalization: Accelerating deep network
  training by reducing internal covariate shift}. In
  \bibinfo{booktitle}{\emph{International conference on machine learning}}.
  PMLR, \bibinfo{pages}{448--456}.
\newblock


\bibitem[\protect\citeauthoryear{Kahn, Villaflor, Pong, Abbeel, and
  Levine}{Kahn et~al\mbox{.}}{2017}]%
        {kahn2017uncertainty}
\bibfield{author}{\bibinfo{person}{Gregory Kahn}, \bibinfo{person}{Adam
  Villaflor}, \bibinfo{person}{Vitchyr Pong}, \bibinfo{person}{Pieter Abbeel},
  {and} \bibinfo{person}{Sergey Levine}.} \bibinfo{year}{2017}\natexlab{}.
\newblock \showarticletitle{Uncertainty-aware reinforcement learning for
  collision avoidance}.
\newblock \bibinfo{journal}{\emph{arXiv preprint arXiv:1702.01182}}
  (\bibinfo{year}{2017}).
\newblock


\bibitem[\protect\citeauthoryear{Kendall, Badrinarayanan, and Cipolla}{Kendall
  et~al\mbox{.}}{2015}]%
        {kendall2015bayesian}
\bibfield{author}{\bibinfo{person}{Alex Kendall}, \bibinfo{person}{Vijay
  Badrinarayanan}, {and} \bibinfo{person}{Roberto Cipolla}.}
  \bibinfo{year}{2015}\natexlab{}.
\newblock \showarticletitle{Bayesian segnet: Model uncertainty in deep
  convolutional encoder-decoder architectures for scene understanding}.
\newblock \bibinfo{journal}{\emph{arXiv preprint arXiv:1511.02680}}
  (\bibinfo{year}{2015}).
\newblock


\bibitem[\protect\citeauthoryear{Kingma and Welling}{Kingma and
  Welling}{2013}]%
        {kingma2013auto}
\bibfield{author}{\bibinfo{person}{Diederik~P Kingma} {and}
  \bibinfo{person}{Max Welling}.} \bibinfo{year}{2013}\natexlab{}.
\newblock \showarticletitle{Auto-encoding variational bayes}.
\newblock \bibinfo{journal}{\emph{arXiv preprint arXiv:1312.6114}}
  (\bibinfo{year}{2013}).
\newblock


\bibitem[\protect\citeauthoryear{Lakshminarayanan, Pritzel, and
  Blundell}{Lakshminarayanan et~al\mbox{.}}{2017}]%
        {lakshminarayanan2017simple}
\bibfield{author}{\bibinfo{person}{Balaji Lakshminarayanan},
  \bibinfo{person}{Alexander Pritzel}, {and} \bibinfo{person}{Charles
  Blundell}.} \bibinfo{year}{2017}\natexlab{}.
\newblock \showarticletitle{Simple and scalable predictive uncertainty
  estimation using deep ensembles}. In \bibinfo{booktitle}{\emph{Advances in
  neural information processing systems}}. \bibinfo{pages}{6402--6413}.
\newblock


\bibitem[\protect\citeauthoryear{Laptev, Yosinski, Li, and Smyl}{Laptev
  et~al\mbox{.}}{2017}]%
        {laptev2017time}
\bibfield{author}{\bibinfo{person}{Nikolay Laptev}, \bibinfo{person}{Jason
  Yosinski}, \bibinfo{person}{Li~Erran Li}, {and} \bibinfo{person}{Slawek
  Smyl}.} \bibinfo{year}{2017}\natexlab{}.
\newblock \showarticletitle{Time-series extreme event forecasting with neural
  networks at uber}. In \bibinfo{booktitle}{\emph{International conference on
  machine learning}}, Vol.~\bibinfo{volume}{34}. \bibinfo{pages}{1--5}.
\newblock


\bibitem[\protect\citeauthoryear{Laves, Ihler, Ortmaier, and Kahrs}{Laves
  et~al\mbox{.}}{2019}]%
        {laves2019quantifying}
\bibfield{author}{\bibinfo{person}{Max-Heinrich Laves}, \bibinfo{person}{Sontje
  Ihler}, \bibinfo{person}{Tobias Ortmaier}, {and} \bibinfo{person}{L{\"u}der~A
  Kahrs}.} \bibinfo{year}{2019}\natexlab{}.
\newblock \showarticletitle{Quantifying the uncertainty of deep learning-based
  computer-aided diagnosis for patient safety}.
\newblock \bibinfo{journal}{\emph{Current Directions in Biomedical
  Engineering}} \bibinfo{volume}{5}, \bibinfo{number}{1}
  (\bibinfo{year}{2019}), \bibinfo{pages}{223--226}.
\newblock


\bibitem[\protect\citeauthoryear{LeCun, Bottou, Bengio, and Haffner}{LeCun
  et~al\mbox{.}}{1998}]%
        {lecun1998gradient}
\bibfield{author}{\bibinfo{person}{Yann LeCun}, \bibinfo{person}{L{\'e}on
  Bottou}, \bibinfo{person}{Yoshua Bengio}, {and} \bibinfo{person}{Patrick
  Haffner}.} \bibinfo{year}{1998}\natexlab{}.
\newblock \showarticletitle{Gradient-based learning applied to document
  recognition}.
\newblock \bibinfo{journal}{\emph{Proc. IEEE}} \bibinfo{volume}{86},
  \bibinfo{number}{11} (\bibinfo{year}{1998}), \bibinfo{pages}{2278--2324}.
\newblock


\bibitem[\protect\citeauthoryear{Malinin, Mlodozeniec, and Gales}{Malinin
  et~al\mbox{.}}{2019}]%
        {malinin2019ensemble}
\bibfield{author}{\bibinfo{person}{Andrey Malinin}, \bibinfo{person}{Bruno
  Mlodozeniec}, {and} \bibinfo{person}{Mark Gales}.}
  \bibinfo{year}{2019}\natexlab{}.
\newblock \showarticletitle{Ensemble distribution distillation}.
\newblock \bibinfo{journal}{\emph{arXiv preprint arXiv:1905.00076}}
  (\bibinfo{year}{2019}).
\newblock


\bibitem[\protect\citeauthoryear{Mariet, Jenatton, Wenzel, and Tran}{Mariet
  et~al\mbox{.}}{2020}]%
        {mariet2020distilling}
\bibfield{author}{\bibinfo{person}{Zelda~E Mariet}, \bibinfo{person}{Rodolphe
  Jenatton}, \bibinfo{person}{Florian Wenzel}, {and} \bibinfo{person}{Dustin
  Tran}.} \bibinfo{year}{2020}\natexlab{}.
\newblock \showarticletitle{Distilling ensembles improves uncertainty
  estimates}. In \bibinfo{booktitle}{\emph{Third Symposium on Advances in
  Approximate Bayesian Inference}}.
\newblock


\bibitem[\protect\citeauthoryear{Mobiny, Yuan, Moulik, Garg, Wu, and
  Van~Nguyen}{Mobiny et~al\mbox{.}}{2021}]%
        {mobiny2021dropconnect}
\bibfield{author}{\bibinfo{person}{Aryan Mobiny}, \bibinfo{person}{Pengyu
  Yuan}, \bibinfo{person}{Supratik~K Moulik}, \bibinfo{person}{Naveen Garg},
  \bibinfo{person}{Carol~C Wu}, {and} \bibinfo{person}{Hien Van~Nguyen}.}
  \bibinfo{year}{2021}\natexlab{}.
\newblock \showarticletitle{Dropconnect is effective in modeling uncertainty of
  bayesian deep networks}.
\newblock \bibinfo{journal}{\emph{Scientific reports}} \bibinfo{volume}{11},
  \bibinfo{number}{1} (\bibinfo{year}{2021}), \bibinfo{pages}{1--14}.
\newblock


\bibitem[\protect\citeauthoryear{Mukhoti and Gal}{Mukhoti and Gal}{2018}]%
        {mukhoti2018evaluating}
\bibfield{author}{\bibinfo{person}{Jishnu Mukhoti} {and} \bibinfo{person}{Yarin
  Gal}.} \bibinfo{year}{2018}\natexlab{}.
\newblock \showarticletitle{Evaluating bayesian deep learning methods for
  semantic segmentation}.
\newblock \bibinfo{journal}{\emph{arXiv preprint arXiv:1811.12709}}
  (\bibinfo{year}{2018}).
\newblock


\bibitem[\protect\citeauthoryear{Nair and Hinton}{Nair and Hinton}{2010}]%
        {nair2010rectified}
\bibfield{author}{\bibinfo{person}{Vinod Nair} {and}
  \bibinfo{person}{Geoffrey~E Hinton}.} \bibinfo{year}{2010}\natexlab{}.
\newblock \showarticletitle{Rectified linear units improve restricted boltzmann
  machines}. In \bibinfo{booktitle}{\emph{ICML}}.
\newblock


\bibitem[\protect\citeauthoryear{Ovadia, Fertig, Ren, Nado, Sculley, Nowozin,
  Dillon, Lakshminarayanan, and Snoek}{Ovadia et~al\mbox{.}}{2019}]%
        {ovadia2019can}
\bibfield{author}{\bibinfo{person}{Yaniv Ovadia}, \bibinfo{person}{Emily
  Fertig}, \bibinfo{person}{Jie Ren}, \bibinfo{person}{Zachary Nado},
  \bibinfo{person}{David Sculley}, \bibinfo{person}{Sebastian Nowozin},
  \bibinfo{person}{Joshua Dillon}, \bibinfo{person}{Balaji Lakshminarayanan},
  {and} \bibinfo{person}{Jasper Snoek}.} \bibinfo{year}{2019}\natexlab{}.
\newblock \showarticletitle{Can you trust your model's uncertainty? Evaluating
  predictive uncertainty under dataset shift}. In
  \bibinfo{booktitle}{\emph{Advances in Neural Information Processing
  Systems}}. \bibinfo{pages}{13991--14002}.
\newblock


\bibitem[\protect\citeauthoryear{Paisley, Blei, and Jordan}{Paisley
  et~al\mbox{.}}{2012}]%
        {paisley2012variational}
\bibfield{author}{\bibinfo{person}{John Paisley}, \bibinfo{person}{David Blei},
  {and} \bibinfo{person}{Michael Jordan}.} \bibinfo{year}{2012}\natexlab{}.
\newblock \showarticletitle{Variational Bayesian inference with stochastic
  search}.
\newblock \bibinfo{journal}{\emph{arXiv preprint arXiv:1206.6430}}
  (\bibinfo{year}{2012}).
\newblock


\bibitem[\protect\citeauthoryear{Papamarkou, Hinkle, Young, and
  Womble}{Papamarkou et~al\mbox{.}}{2019}]%
        {papamarkou2019challenges}
\bibfield{author}{\bibinfo{person}{Theodore Papamarkou}, \bibinfo{person}{Jacob
  Hinkle}, \bibinfo{person}{M~Todd Young}, {and} \bibinfo{person}{David
  Womble}.} \bibinfo{year}{2019}\natexlab{}.
\newblock \showarticletitle{Challenges in Markov chain Monte Carlo for Bayesian
  neural networks}.
\newblock \bibinfo{journal}{\emph{arXiv preprint arXiv:1910.06539}}
  (\bibinfo{year}{2019}).
\newblock


\bibitem[\protect\citeauthoryear{Shamir and Coviello}{Shamir and
  Coviello}{2020}]%
        {shamir20anti}
\bibfield{author}{\bibinfo{person}{Gil~I Shamir} {and} \bibinfo{person}{Lorenzo
  Coviello}.} \bibinfo{year}{2020}\natexlab{}.
\newblock \showarticletitle{Anti-Distillation: Improving Reproducibility of
  Deep Networks}.
\newblock \bibinfo{journal}{\emph{arXiv preprint arXiv:2010.09923}}
  (\bibinfo{year}{2020}).
\newblock


\bibitem[\protect\citeauthoryear{Shamir, Lin, and Coviello}{Shamir
  et~al\mbox{.}}{2020}]%
        {shamir20smooth}
\bibfield{author}{\bibinfo{person}{Gil~I Shamir}, \bibinfo{person}{Dong Lin},
  {and} \bibinfo{person}{Lorenzo Coviello}.} \bibinfo{year}{2020}\natexlab{}.
\newblock \showarticletitle{Smooth activations and reproducibility in deep
  networks}.
\newblock \bibinfo{journal}{\emph{arXiv preprint arXiv:2010.09931}}
  (\bibinfo{year}{2020}).
\newblock


\bibitem[\protect\citeauthoryear{Snapp and Shamir}{Snapp and Shamir}{2021}]%
        {snapp2021synthesizing}
\bibfield{author}{\bibinfo{person}{Robert~R Snapp} {and} \bibinfo{person}{Gil~I
  Shamir}.} \bibinfo{year}{2021}\natexlab{}.
\newblock \showarticletitle{Synthesizing Irreproducibility in Deep Networks}.
\newblock \bibinfo{journal}{\emph{arXiv preprint arXiv:2102.10696}}
  (\bibinfo{year}{2021}).
\newblock


\bibitem[\protect\citeauthoryear{Srivastava, Hinton, Krizhevsky, Sutskever, and
  Salakhutdinov}{Srivastava et~al\mbox{.}}{2014}]%
        {srivastava2014dropout}
\bibfield{author}{\bibinfo{person}{Nitish Srivastava},
  \bibinfo{person}{Geoffrey Hinton}, \bibinfo{person}{Alex Krizhevsky},
  \bibinfo{person}{Ilya Sutskever}, {and} \bibinfo{person}{Ruslan
  Salakhutdinov}.} \bibinfo{year}{2014}\natexlab{}.
\newblock \showarticletitle{Dropout: a simple way to prevent neural networks
  from overfitting}.
\newblock \bibinfo{journal}{\emph{The journal of machine learning research}}
  \bibinfo{volume}{15}, \bibinfo{number}{1} (\bibinfo{year}{2014}),
  \bibinfo{pages}{1929--1958}.
\newblock


\bibitem[\protect\citeauthoryear{Summers and Dinneen}{Summers and
  Dinneen}{2021}]%
        {summers2021nondeterminism}
\bibfield{author}{\bibinfo{person}{Cecilia Summers} {and}
  \bibinfo{person}{Michael~J Dinneen}.} \bibinfo{year}{2021}\natexlab{}.
\newblock \showarticletitle{Nondeterminism and Instability in Neural Network
  Optimization}.
\newblock \bibinfo{journal}{\emph{arXiv preprint arXiv:2103.04514}}
  (\bibinfo{year}{2021}).
\newblock


\bibitem[\protect\citeauthoryear{Tishby and Zaslavsky}{Tishby and
  Zaslavsky}{2015}]%
        {tishby2015deep}
\bibfield{author}{\bibinfo{person}{Naftali Tishby} {and} \bibinfo{person}{Noga
  Zaslavsky}.} \bibinfo{year}{2015}\natexlab{}.
\newblock \showarticletitle{Deep learning and the information bottleneck
  principle}. In \bibinfo{booktitle}{\emph{2015 IEEE Information Theory
  Workshop (ITW)}}. IEEE, \bibinfo{pages}{1--5}.
\newblock


\bibitem[\protect\citeauthoryear{Valdenegro-Toro}{Valdenegro-Toro}{2019}]%
        {valdenegro2019deep}
\bibfield{author}{\bibinfo{person}{Matias Valdenegro-Toro}.}
  \bibinfo{year}{2019}\natexlab{}.
\newblock \showarticletitle{Deep sub-ensembles for fast uncertainty estimation
  in image classification}.
\newblock \bibinfo{journal}{\emph{arXiv preprint arXiv:1910.08168}}
  (\bibinfo{year}{2019}).
\newblock


\bibitem[\protect\citeauthoryear{Wan, Zeiler, Zhang, Le~Cun, and Fergus}{Wan
  et~al\mbox{.}}{2013}]%
        {wan2013regularization}
\bibfield{author}{\bibinfo{person}{Li Wan}, \bibinfo{person}{Matthew Zeiler},
  \bibinfo{person}{Sixin Zhang}, \bibinfo{person}{Yann Le~Cun}, {and}
  \bibinfo{person}{Rob Fergus}.} \bibinfo{year}{2013}\natexlab{}.
\newblock \showarticletitle{Regularization of neural networks using
  dropconnect}. In \bibinfo{booktitle}{\emph{International conference on
  machine learning}}. PMLR, \bibinfo{pages}{1058--1066}.
\newblock


\bibitem[\protect\citeauthoryear{Wen and Tadmor}{Wen and Tadmor}{2020}]%
        {wen2020uncertainty}
\bibfield{author}{\bibinfo{person}{Mingjian Wen} {and} \bibinfo{person}{Ellad~B
  Tadmor}.} \bibinfo{year}{2020}\natexlab{}.
\newblock \showarticletitle{Uncertainty quantification in molecular simulations
  with dropout neural network potentials}.
\newblock \bibinfo{journal}{\emph{npj Computational Materials}}
  \bibinfo{volume}{6}, \bibinfo{number}{1} (\bibinfo{year}{2020}),
  \bibinfo{pages}{1--10}.
\newblock


\bibitem[\protect\citeauthoryear{Wen, Tran, and Ba}{Wen et~al\mbox{.}}{2020}]%
        {wen2020batchens}
\bibfield{author}{\bibinfo{person}{Yeming Wen}, \bibinfo{person}{Dustin Tran},
  {and} \bibinfo{person}{Jimmy Ba}.} \bibinfo{year}{2020}\natexlab{}.
\newblock \showarticletitle{BatchEnsemble: An Alternative Approach to Efficient
  Ensemble and Lifelong Learning}.
\newblock \bibinfo{journal}{\emph{Eighth International Conference on Learning
  Representations (ICLR 2020)}} (\bibinfo{year}{2020}).
\newblock


\bibitem[\protect\citeauthoryear{Wilson and Izmailov}{Wilson and
  Izmailov}{2020}]%
        {wilson2020bayesian}
\bibfield{author}{\bibinfo{person}{Andrew~Gordon Wilson} {and}
  \bibinfo{person}{Pavel Izmailov}.} \bibinfo{year}{2020}\natexlab{}.
\newblock \showarticletitle{Bayesian deep learning and a probabilistic
  perspective of generalization}.
\newblock \bibinfo{journal}{\emph{arXiv preprint arXiv:2002.08791}}
  (\bibinfo{year}{2020}).
\newblock


\bibitem[\protect\citeauthoryear{Yang, Ma, Nie, Chang, and Hauptmann}{Yang
  et~al\mbox{.}}{2015}]%
        {yang2015multi}
\bibfield{author}{\bibinfo{person}{Yi Yang}, \bibinfo{person}{Zhigang Ma},
  \bibinfo{person}{Feiping Nie}, \bibinfo{person}{Xiaojun Chang}, {and}
  \bibinfo{person}{Alexander~G Hauptmann}.} \bibinfo{year}{2015}\natexlab{}.
\newblock \showarticletitle{Multi-class active learning by uncertainty sampling
  with diversity maximization}.
\newblock \bibinfo{journal}{\emph{International Journal of Computer Vision}}
  \bibinfo{volume}{113}, \bibinfo{number}{2} (\bibinfo{year}{2015}),
  \bibinfo{pages}{113--127}.
\newblock


\bibitem[\protect\citeauthoryear{Yao, Yi, Cheng, Yu, Chen, Menon, Hong, Chi,
  Tjoa, Kang, et~al\mbox{.}}{Yao et~al\mbox{.}}{2021}]%
        {yao2020self}
\bibfield{author}{\bibinfo{person}{Tiansheng Yao}, \bibinfo{person}{Xinyang
  Yi}, \bibinfo{person}{Derek~Zhiyuan Cheng}, \bibinfo{person}{Felix Yu},
  \bibinfo{person}{Ting Chen}, \bibinfo{person}{Aditya Menon},
  \bibinfo{person}{Lichan Hong}, \bibinfo{person}{Ed~H Chi},
  \bibinfo{person}{Steve Tjoa}, \bibinfo{person}{Jieqi Kang}, {et~al\mbox{.}}}
  \bibinfo{year}{2021}\natexlab{}.
\newblock \showarticletitle{Self-supervised Learning for Large-scale Item
  Recommendations}.
\newblock \bibinfo{journal}{\emph{CIKM}} (\bibinfo{year}{2021}).
\newblock


\bibitem[\protect\citeauthoryear{Zhang, Bengio, and Singer}{Zhang
  et~al\mbox{.}}{2019}]%
        {zhang2019all}
\bibfield{author}{\bibinfo{person}{Chiyuan Zhang}, \bibinfo{person}{Samy
  Bengio}, {and} \bibinfo{person}{Yoram Singer}.}
  \bibinfo{year}{2019}\natexlab{}.
\newblock \showarticletitle{Are all layers created equal?}
\newblock \bibinfo{journal}{\emph{arXiv preprint arXiv:1902.01996}}
  (\bibinfo{year}{2019}).
\newblock


\end{thebibliography}

\end{document}